\newcommand{\vecpn}{{\mathbf{p}_n}}
\newcommand{\vecpb}{{\mathbf{p}_b}}
\newcommand{\vecpm}{{\mathbf{p}_m}}
\newcommand{\vecpnu}{{p_n^u}}
\newcommand{\vecpnv}{{p_n^v}}
\newcommand{\mloss}[1]{\mathcal{L}_{main_#1}}
\newcommand{\aloss}[1]{\mathcal{L}_{aux_#1}}

\newcommand{\ts}[4]{$#1$ & $#2$ & $#3$ & $#4$}
\newcommand{\tc}[9]{$#1$ & $#2$ & $#3$ & $#4$ & $#5$ & $#6$ & $#7$ & $#8$ & $#9$}
\newcommand{\TC}[9]{\boldmath{$#1$} & \boldmath{$#2$} & \boldmath{$#3$} & \boldmath{$#4$} & \boldmath{$#5$} & \boldmath{$#6$} & \boldmath{$#7$} & \boldmath{$#8$} & \boldmath{$#9$}}

\newcommand{\tabincell}[2]{\begin{tabular}{@{}#1@{}}#2\end{tabular}}

\documentclass[journal, twoside]{IEEEtran}
\usepackage[caption=false,font=footnotesize]{subfig}
\usepackage[pdftex]{graphicx}
\usepackage{graphics} 
\usepackage{epsfig} 
\graphicspath{{images/}{figures/}}
\DeclareGraphicsExtensions{.eps,.PNG,.png, .jpg}
\usepackage{booktabs}
\usepackage{tabu}
\usepackage{xcolor}
\usepackage[normalem]{ulem} 
\newcommand\hl{\bgroup\markoverwith
	{\textcolor{yellow}{\rule[-.5ex]{2pt}{2.5ex}}}\ULon}
\usepackage[switch,columnwise]{lineno}

\usepackage{stfloats} 

\usepackage{cite}

\ifCLASSINFOpdf

\else

\fi

\usepackage{amsmath}

\begin{document}
	\title{Restricted Deformable Convolution based\\Road Scene Semantic Segmentation\\Using Surround View Cameras}	
	\author{Liuyuan Deng, Ming Yang, Hao Li, Tianyi Li, Bing Hu, and Chunxiang Wang
	\thanks{Manuscript received December 17, 2017; revised October 9, 2018, March 10, 2019, and July 18, 2019; accepted August 28, 2019. This work was supported in part by the National Natural Science Foundation of China under Grant U1764264 and Grant 61873165, in part by the Shanghai Automotive Industry Science and Technology Development Foundation under Grant 1733 and Grant 1807, and in part by the International Chair on Automated Driving of Ground Vehicle. The Associate Editor for this article was D. Fernandez-Llorca. (Corresponding author: Ming Yang.)}
		\thanks{L. Deng, M. Yang, T. Li, B. Hu, and C. Wang are with the Department of Automation, Shanghai Jiao Tong University, Shanghai, 200240, and also with the Key Laboratory of System Control and Information Processing, Ministry of Education of China, Shanghai, 200240, China (phone: +86-21-34204553; e-mail: mingyang@sjtu.edu.cn).}
		\thanks{H. Li is with SJTU-ParisTech Elite Institute of Technology and also with Department of Automation, Shanghai Jiao Tong University, Shanghai, 200240, China.}
		\thanks{Digital Object Identifier 10.1109/TITS.2019.2939832}
	}
	
	\markboth{IEEE Transactions on Intelligent Transportation Systems}{Deng \MakeLowercase{\textit{et al.}}: Restricted Deformable Convolution based Semantic Segmentation Using Surround View Cameras}

	\maketitle

	\begin{abstract}
		Understanding the surrounding environment of the vehicle is still one of the challenges for autonomous driving. This paper addresses 360-degree road scene semantic segmentation using surround view cameras, which are widely equipped in existing production cars. First, in order to address large distortion problem in the fisheye images, Restricted Deformable Convolution (RDC) is proposed for semantic segmentation, which can effectively model geometric transformations by learning the shapes of convolutional filters conditioned on the input feature map. 
		Second, in order to obtain a large-scale training set of surround view images, a novel method called zoom augmentation is proposed to transform conventional images to fisheye images. 
		Finally, an RDC based semantic segmentation model is built; the model is trained for real-world surround view images through a multi-task learning architecture by combining real-world images with transformed images. Experiments demonstrate the effectiveness of the RDC to handle images with large distortions, and that the proposed approach shows a good performance using surround view cameras with the help of the transformed images.
		
	\end{abstract}
	
	\begin{IEEEkeywords}
		Deformable convolution, semantic segmentation, road scene understanding, surround view cameras, multi-task learning.
	\end{IEEEkeywords}

	\IEEEpeerreviewmaketitle

	\section{Introduction}
	
	\IEEEPARstart{A}{utonomous} vehicles need to perceive and understand their surroundings (such as road users, free space, and other road scene semantics) for decision making, path planning, etc. Since Nissan introduced the surround view camera system in 2007 on the Infiniti EX35, many Tier1s and OEMs are actively developing such technology. Besides Infiniti and Nissan, automakers such as BMW, Audi, Mercedes Benz, Lexus, and Toyota offer similar systems in their production vehicles. The system usually consists of four fisheye cameras mounted around the vehicle to provide 360-degree surroundings, which helps eliminate all blind spots during critical and precise maneuvers. Based on the surround view cameras, this paper explores the 360-degree road scene understanding.
	
		\begin{figure}[!t]
		\centering
		\subfloat[Raw fisheye image]{\includegraphics[width=0.46\columnwidth]{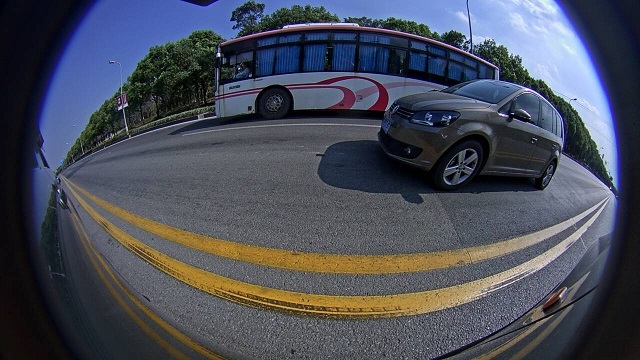}
		}
		\subfloat[Undistorted image]{\includegraphics[width=0.46\columnwidth]{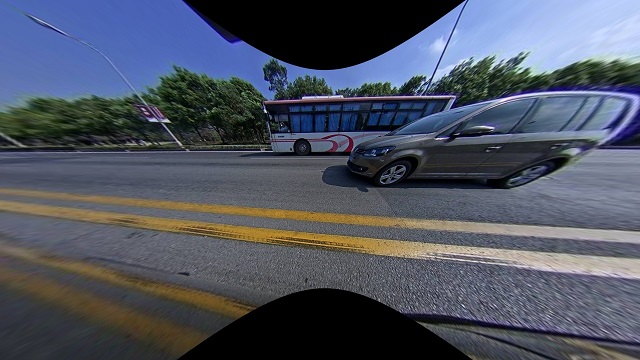}
		}
		\caption{Illustration of image undistortion. The center of the undistorted image is clear, but the boundaries of the image are very blurred. And some information is lost during transferring the pixels of the raw fisheye image into the undistorted image.}
		\label{distortion}
	\end{figure}
	\begin{figure}[!t]
		\centering
		\includegraphics[width=0.95\columnwidth]{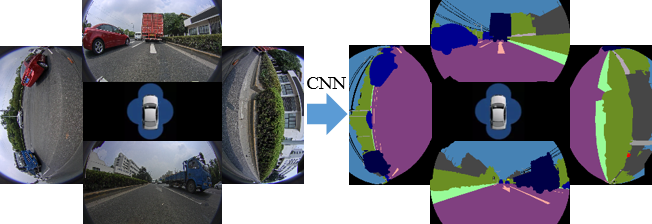}
		\caption{Illustration of CNN based semantic segmentation on raw surround view images. Surround view cameras consist of four fisheye cameras mounted on each side of the vehicle. Cameras in different directions capture images with different image composition.}
		\label{fig: Fig. 1}
	\end{figure}
	
	Thanks to the methodology of Convolutional Neural Network (CNN) based semantic segmentation, in recent years, road scene understanding has achieved huge progress using narrow-angle or even wide-angle conventional cameras\cite{siam2017deep}. Conventional cameras follow well the pinhole camera model: all straight lines in the real world are projected as straight lines in the image. However, images captured by fisheye cameras have strong distortions. As distortions bring difficulties in image processing, fisheye images are usually undistorted before use\cite{wang2014automatic, varga2017super}. However, image undistortion hurts image quality (especially at image boundaries)\cite{fremont2016vision} and leads to information loss. An example of image undistortion is shown in Fig.~\ref{distortion}. We consider that the segmented results on raw fisheye images can be used as an information source for other tasks. For example, visual odometry system for fisheye cameras \cite{seok2019robust} can use the semantics to improve the performance like visual semantic odometry (VSO)\cite{lianos2018vso}. This paper explores CNN based semantic segmentation on raw surround view images, as illustrated in Fig.~\ref{fig: Fig. 1}.
	
	\newpage
	Two challenging aspects are considered. The first is an effective deep learning model to handle fisheye images. Fisheye images have severe distortion which is unavoidable during the process of projecting an image of a hemispheric field onto a plane\cite{miyamoto1964fish}. The degree of distortion is related to the distance between the camera and the objects, and also to the radial angle. The distortions are not uniform over all spatial areas\cite{fremont2016vision}. This brings CNN models the demand for modeling large and unknown transformations. CNNs already have shown a remarkable representing ability to handle distortions with the help of large-scale datasets which contain diverse scenes. The ability largely originates from the large capacity of deep models like VGGNet\cite{simonyan2014very}, GoogleNet\cite{szegedy2015going} and ResNet\cite{he2016deep}. Besides, handcrafted structures, for example, pyramid pooling module\cite{zhao2016pyramid}, also contribute to the representational power. However, regular CNNs have inherently limited ability to model large, unknown geometric distortions\cite{dai2017deformable}. The CNNs have fixed structures, such as fixed filter kernels, fixed receptive field sizes, and fixed pooling kernels. There lack internal mechanisms to handle the geometric distortions. Interested readers may refer to \cite{dai2017deformable} for details.
	
	The second is about training datasets for the deep neural networks. So far, state-of-the-art CNN-based semantic segmentation methods require large-scale pixel-level annotated images for parameter optimization. The annotating process is time-consuming and expensive work, yet several road scene datasets have already been created\cite{brostow2009semantic,cordts2016cityscapes} and contribute to the development of semantic segmentation algorithms. However, there are few large-scale annotated datasets of semantic segmentation for surround view cameras. In our previous works\cite{deng2017cnn}, a fisheye dataset is generated from Cityscapes dataset for a forward-looking conventional camera. However, it is not enough for surround view cameras. First, the image composition of cameras in different directions varies a lot. For example, as shown in Fig.~\ref{fig: Fig. 1}, a forward-looking camera usually captures the rear view of front vehicles, but a sideways-looking camera captures the side view of surround vehicles. Second, the Cityscapes dataset is collected from cities in Europe; the model trained using such dataset may not be suitable for applications in regions outside Europe.

	This paper is a considerable extension of our previous conference publication\cite{deng2017cnn}. We further address the method of road scene semantic segmentation using surround view cameras with a more comprehensive set of improvements and experiments. The main contributions w.r.t. the previous work are as follows.
	
	First, a more effective module is proposed to handle images with large distortions. We do not use the OPP module proposed in \cite{deng2017cnn} because it does not show improvements with the highly efficient ERFNet. Instead, we explore the deformable convolution\cite{dai2017deformable} to handle fisheye images. To address the spatial correspondence problem\cite{li2017dense}, the Restricted Deformable Convolution (RDC) is proposed to further restrict deformable convolution for pixel-wise prediction tasks.
	
	Second, zoom augmentation is redefined as the operation of transforming existing conventional images to fisheye images, and a zoom augmentation layer with a CUDA implementation is implemented for online training. Conventional datasets are used to augment the surround view images via the zoom augmentation method.
	
	Finally, a multi-task learning architecture is presented to train an end-to-end semantic segmentation model for real-world surround view images by combining a small number of real-world images and a large number of transformed images. We introduce the idea of AdaBN\cite{li2016revisiting} to bridge the distributional gap between real-world images and transformed images. In addition, the Hybrid Loss Weightings (HLW) is proposed to improve the generalization ability by introducing auxiliary losses with different loss weightings.
	
	This paper is organized as follows: Section II reviews related works. Section III introduces the RDC, whereas Section IV describes the method of converting the existing datasets to fisheye datasets. Section V presents the training strategy. And Section VI demonstrates quantitative experiments.
	
	\section{Related Work}
	Early semantic segmentation methods rely on handcrafted features; they use Random Decision Forest\cite{scharwachter2015low} or Boosting\cite{sturgess2009combining} to predict the class probabilities and use probabilistic models known as Conditional Random Fields (CRFs) to handle uncertainties and propagate contextual information across the image. In recent years, CNNs have made a huge step forward in vision recognition thanks to large-scale training datasets and high-performance Graphics Processing Unit (GPU). In addition, excellent open-source deep learning frameworks like Caffe, MXNet, and Tensorflow boost the development of algorithms. Powerful deep neural networks\cite{simonyan2014very,szegedy2015going,he2016deep} emerged largely reducing the classification errors on ImageNet\cite{russakovsky2015imagenet}, which is also beneficial to semantic segmentation. FCN\cite{shelhamer2017fully} successfully improved the accuracy of semantic segmentation by adapting classification networks into fully convolutional networks.
	
	For the task of semantic segmentation, it is crucial to incorporate context information in relevant image regions when making a prediction. A broad receptive field is usually desirable to capture the entire useful information. The receptive field size can be increased multiplicatively by down-sampling operation and linearly by stacking more layers. After the down-sampling operation, lots of low-level visual features are lost, and the spatial structure of the scene is prone to be damaged. Dilated convolution or Atrous convolution\cite{yu2015multi,chen2016deeplab} is proposed to alleviate this problem by enlarging the receptive field without reducing the spatial resolution. It enlarges the kernel size by introducing “holes” in convolution filters without increasing the number of parameters. Note that the dilation rates should be carefully designed to alleviate gridding artifacts\cite{wang2017understanding}. On the other hand, modern nets like ResNet\cite{he2016deep} theoretically have a large receptive field, even larger than the input image, due to the significantly increased depth. However, as investigated in\cite{zhou2014object}, the effective receptive field of a network is much smaller than the theoretical one. Instead of hand-crafted designing modules, the deformable convolution\cite{dai2017deformable} learns the shapes of convolution filters conditioned on an input feature map. The receptive field and the spatial sampling locations are adapted according to the objects' scale and shape. It is shown that it is feasible and effective to learn geometric transformation in CNNs for vision tasks. However, as indicated in\cite{li2017dense}, the deformable convolution does not address the spatial correspondence problem which is critical in dense prediction tasks. The DTN\cite{li2017dense} preserves the spatial correspondence of spatial transformer layers between the input and output and uses a corresponding decoder layer to restore the correspondence. However, the DTN learns a global parametric transformation, which is limited to model non-uniform geometric transformations for each location.
	
	Some datasets for semantic road scene understanding have been created, for example, CamVid\cite{brostow2009semantic}, Cityscapes\cite{cordts2016cityscapes}, and Mapillary Vistas\cite{neuhold2017mapillary}. 
	Cityscapes is a large-scale dataset for semantic urban scene understanding with 5000 finely annotated images. The images are captured from Europe using forward-looking conventional cameras. The pixel-level data annotation as well as data collection is time-consuming and expensive.
	Alvarez et al. \cite{alvarez2012road} addressed the problem by using an algorithm trained on a general image dataset to generate noisy labels on unseen images. Then the noisy labels were used to guide the training for road scene segmentation.
	Another increasingly popular way to overcome the lack of large-scale dataset is explored by the usage of synthetic data, such as VEIS\cite{saleh2018effective}, SYNTHIA\cite{ros2016synthia}, Virtual KITTI\cite{gaidon2016virtual}, and GTA-V\cite{richter2016playing}. Synthetic data is usually used to augment real training data\cite{ros2016synthia,de2016procedural}. The SYNTHIA dataset is generated by rendering a virtual city created with the Unity development platform for semantic segmentation of driving scenes. Saleh et al. \cite{saleh2018effective} proposed VEIS environment to generate the VEIS dataset which has richer foreground classes of real traffic environments. Our previous works\cite{deng2017cnn} can also be regarded as a synthetic dataset which is transformed from a real large-scale conventional image dataset. None of these datasets is created using surround view cameras.
	
	To overcome the burden of annotation, weakly-supervised methods using a weaker form of annotation such as image tags and bounding boxes, and domain adaptation methods using annotated data only in source domains, have been investigated\cite{saleh2018effective}. Saleh et al.\cite{saleh2018incorporating} explored weakly-supervised semantic segmentation using only image tags and networks pre-trained for the task of object recognition on ImageNet. \cite{saleh2018effective} used the synthetic data only for semantic segmentation of real images. For better performance in local environments, this paper combines a small number of annotated local images and a large number of synthetic data in a multi-task learning architecture.
	
	In order to achieve higher accuracy of semantic segmentation, the top-performing networks based on very large models can be used, e.g., \cite{zhao2016pyramid,wang2017understanding}; however, such methods suffer from high computational costs. As a matter of fact, autonomous driving features multitasking with limited resource. The long inference times and large power consumption make them difficult to be employed in on-road applications. On the other hand, some works\cite{romera2017erfnet,paszke2016enet,treml2016speeding} explore a good tradeoff between accuracy and efficiency, so that it is feasible on embedded devices. ERFNet\cite{romera2017erfnet} achieved an excellent tradeoff by applying factorized convolutions\cite{alvarez2016decomposeme}. Moreover, it can be trained from scratch. This paper takes ERFNet as the baseline model for efficient semantic segmentation.
	
	\begin{figure}[!t]
		\centering	\subfloat[]{\includegraphics[width=0.1\textwidth]{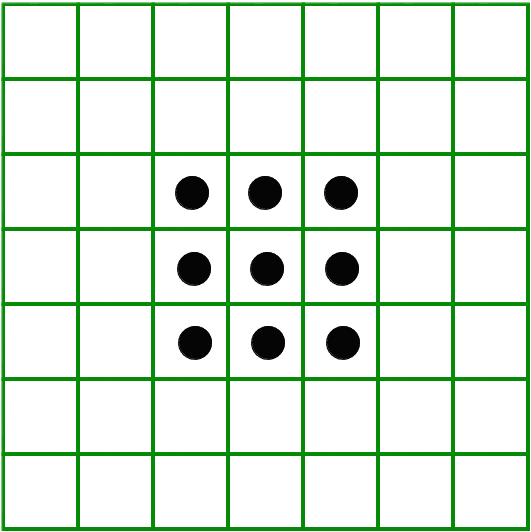}
			\label{fig: Fig. 2(a)}}
		\subfloat[]{\includegraphics[width=0.1\textwidth]{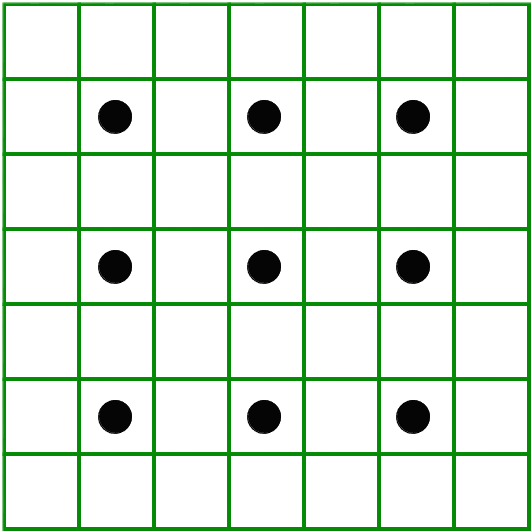}
			\label{fig: Fig. 2(b)}}
		\subfloat[]{\includegraphics[width=0.1\textwidth]{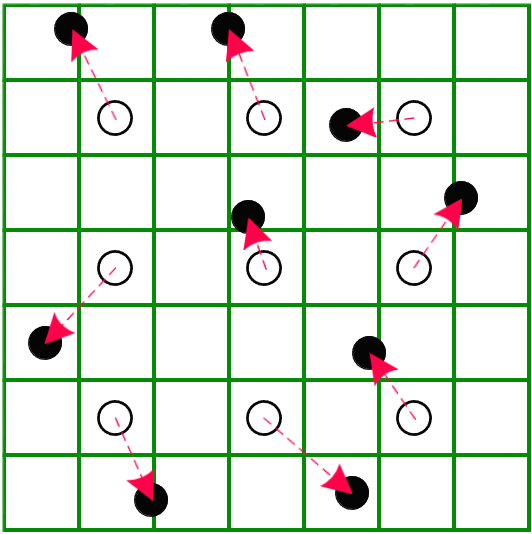}
			\label{fig: Fig. 2(c)}}
		\subfloat[]{\includegraphics[width=0.1\textwidth]{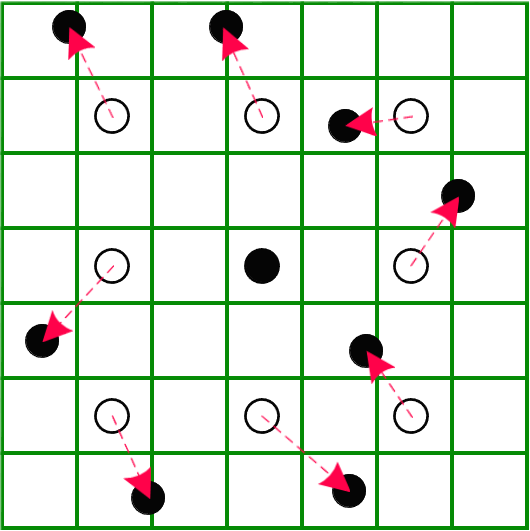}
			\label{fig: Fig. 2(d)}}
		\caption{The sampling locations of 3x3 convolutions: (a) Standard convolution. (b) Dilated convolution with dilation 2. (c) Deformable convolution. (d) Restricted deformable convolution. The dark points are the actual sampling locations, and the hollow circles in (c) and (d) are the initial sampling locations. (a) and (b) employ a fixed grid of sampling locations. (c) and (d) augment the sampling locations with learned 2D offsets (red arrows). The primary difference between (c) and (d) is that restricted deformable convolution employs a fixed central sampling location. No offsets are needed to be learned for the central sampling location in (d).}
		\label{fig: Fig. 2}
	\end{figure}
	\section{Restricted Deformable Convolution}
	In this section, we describe the RDC which is the restricted version of deformable convolution. And a factorized version of RDC is also provided.
	
	The regular convolution adopts a fixed filter with grid sampling locations, as shown in Fig.~\ref{fig: Fig. 2(a)} and Fig.~\ref{fig: Fig. 2(b)}. The shape of a regular grid is a rectangle, for example, as shown in Fig.~\ref{fig: Fig. 2(b)}, a $3\times3$ filter with dilation 2 is defined as:
	$$\mathcal{R}=\{(-2,-2),(0,-2),\ldots,(0,0),\ldots,(0,2),(2,2)\}.$$
	The deformable convolution adds 2D offsets to the grid sampling locations, as shown in Fig.~\ref{fig: Fig. 2(c)}. Thus, each sampling location is learnable and dynamic.
	
	In a deep CNN, the deeper layers encode high-level semantic information with weak spatial information, including object- or category-level evidence. Features from the middle layers are expected to describe middle-level representations for object parts and retain spatial information. Features from the lower convolution layers encode low-level spatial visual information like edges, corners, circles, etc. The middle layers and lower layers are responsible for learning the spatial structures. If the deformable convolution is applied to the lower or middle layers, the spatial structures are susceptible to fluctuation. The spatial correspondence between input images and output label maps is difficult to be preserved. This is the spatial correspondence problem indicated in\cite{li2017dense}, which is critical in pixel-wise semantic segmentation. Hence, the deformable convolution is only applied to the last few convolution layers, as in the works presented in\cite{dai2017deformable}. 
	
\begin{figure}[!t]
		\centering
		\includegraphics[width=0.40\textwidth]{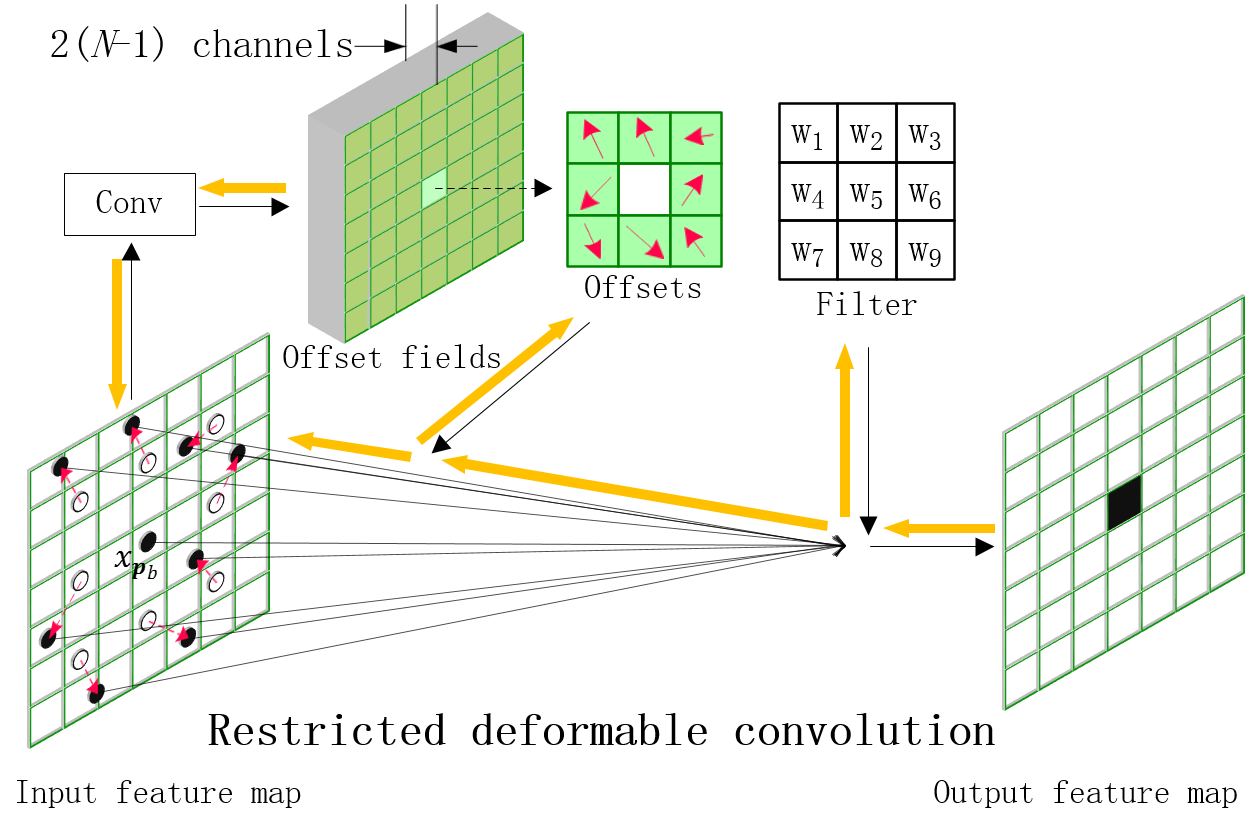}
		\caption{A $3\times3$ restricted deformable convolution. The module is initiated with a $3\times3$ filter with dilation $2$ (the hollow circles on the input feature map). Offset fields are learned from the input feature map by a regular convolutional layer. The channel dimension $2(N-1)$ corresponds $N-1$ 2D offsets (the red arrows). Here, $N=9$. The actual sampling positions (dark points) are obtained by adding the 2D offsets to the initial locations. The value of the new position is obtained by using bilinear interpolation to weight the four nearest points. The yellow arrows denote the backpropagation paths of gradients.}
		\label{fig: Fig. 3}
	\end{figure}

	In this paper, a straightforward way is adopted to alleviate this problem. As illustrated in Fig.~\ref{fig: Fig. 2(d)}, we freeze the central location of the filter and let the outer locations be learnable, considering that the ability of modeling transformations heavily depends on the outer sampling locations. This variant of deformable convolution is called Restricted Deformable Convolution (RDC), as shown in Fig.~\ref{fig: Fig. 3}. The RDC is first initialized with the shape of a filter of regular convolution. Then 2D offsets are learned by a regular convolutional layer to augment the regular grid locations except for the center. The shape of the filter is deformable and learned from the input features. The RDC can be included in a standard neural network architecture to enhance the ability to model geometric transformations. 
	\subsection{Formulation}
	The convolution operator slides a filter or kernel over the input feature map $\mathbf{X}$ to produce output feature map $\mathbf{Y}$. For simplicity, we consider the case with one output channel. For each sliding position $\vecpb$, a regular convolution with filter weights $\mathbf{W}$, bias term $\mathbf{b}$ and stride 1 can be formulated as
	$$\mathbf{Y}=\mathbf{W}*\mathbf{X}+\mathbf{b}$$
	\begin{equation}
	\label{eqn:rconv}
	y_\vecpb=\sum_c{\sum_{\vecpn\in\mathcal{R}}{w_{c,n}\cdot x_{c,\vecpb+\vecpn}+b}}
	\end{equation}
	where $c$ is the index of the input channel, $\vecpb$ is the base position of the convolution, $n=1,\ldots,N$ with $N=|{\mathcal{R}|}$ and $\vecpn\in\mathcal{R}$ enumerates the locations in the regular grid $\mathcal{R}$. The center of $\mathcal{R}$ is denoted as $\vecpm$ which is always equal to $(0,0)$, under the assumption that both of height and width of the kernel are odd numbers, such as $3\times3$, and $1\times3$. This assumption is suitable for most CNNs. $m$ is the index of the central location in $\mathcal{R}$.
	
	The deformable convolution augments all the sampling locations with learned offsets $\{\Delta\mathbf{p}_n|n=1,\ldots,N\}$. Each offset has a horizontal component and a vertical component. Totally $2N$ offset parameters are required to learn for each sliding position. Equation (\ref{eqn:rconv}) becomes
	\begin{equation}
	\label{eqn:dconv}
	y_\vecpb=\sum_{\vecpn\in\mathcal{R}}{w_n\cdot x_{\mathbf{H}(\vecpn)}+b}
	\end{equation}
	where $\mathbf{H}(\vecpn)=\vecpb+\vecpn+\Delta\vecpn$ is the learned sampling position on input feature map. The input channel $c$ in (\ref{eqn:rconv}) is omitted in (\ref{eqn:dconv}) for notation clarity, because the same operation is applied in every channel.
	
	In order to preserve the spatial structure, we restrict the deformable convolution by fixing its central location. That is to say, the offset $\Delta\vecpm$ is set as $(0,0)$. The center of $\mathcal{R}$, $\vecpm$, is also equal to $(0,0)$, thus the learned position is formulated as
	$$ \mathbf{H}(\vecpn)=
	\begin{cases}
	\vecpb& n=m\\
	\vecpb+\vecpn+\Delta\vecpb& n \neq m
	\end{cases}
	$$
	The RDC can also be formulated by
	$$
	y_\vecpb=w_m\cdot x_\vecpb+\sum_{\vecpn\in\mathcal{R},n\neq m}{w_n\cdot x_{\vecpnu,\vecpnv}+b}
	$$
	where $\vecpnu$ and $\vecpnv$ are horizontal and vertical components of $\mathbf{H}(\vecpn)$. The first term of the formula calculates the weighted value for the fixed central location. The second term calculates the weighted sum for the learned outer locations. The learned outer positions $(\vecpnu,\vecpnv)$ are not integer numbers, because the offsets $\Delta\vecpn$ are real numbers. Bilinear interpolation is used to sample over the input feature map for a fractional position. For the outer locations, the sampled output is formulated by
	$$
	x_{\vecpnu,\vecpnv}=
	\begin{bmatrix}
	1-\Delta\vecpnu \\ \Delta\vecpnu
	\end{bmatrix}^T
	\mathbf{Q}
	\begin{bmatrix}
	1-\Delta\vecpnv \\ \Delta\vecpnv
	\end{bmatrix}
	$$
	where
	\begin{gather*}
	\mathbf{Q}=
	\begin{bmatrix}
	x_{\lfloor\vecpnu\rfloor,\lfloor\vecpnv\rfloor} & x_{\lfloor\vecpnu\rfloor,\lceil\vecpnv\rceil} \\
	x_{\lceil\vecpnu\rceil, \lfloor\vecpnv\rfloor} & 
	x_{\lceil\vecpnu\rceil,\lceil\vecpnv\rceil}
	\end{bmatrix}\\
	\Delta\vecpnu=\vecpnu-\lfloor\vecpnu\rfloor\\ \Delta\vecpnv=\vecpnv-\lfloor\vecpnv\rfloor
	\end{gather*}
	$\mathbf{Q}$ denotes the values of four nearest integer positions on the input feature map $\mathbf{X}$. This bilinear interpolation operation is differentiable, as explained in\cite{dai2017deformable}.
	
	As illustrated in Fig.~\ref{fig: Fig. 3}, like the deformable convolution, the offsets $\{\Delta\vecpn|n=1,\ldots,N,n\neq m\}$ in the RDC are learned with a convolutional layer and from the same input feature map. The spatial resolution of the output offset fields is identical to that of the output feature map. Therefore, for each sliding position, a specific shape of the filter is learned. $2(N-1)$ offset parameters are required to define the new shape, whereas no parameter is required to define the offset of the central location.

	The whole module is differentiable. It can be trained with the standard backpropagation method, allowing for the end-to-end training of the models they are injected in. As illustrated in Fig.~\ref{fig: Fig. 3}, the gradients are passed to filter weights, and also backpropagated to the offsets and input feature map through the bilinear operation.
	
	\subsection{Factorized Restricted Deformable Convolution}
	\begin{figure}[!t]
		\centering
		\subfloat[]{\includegraphics[width=0.14\textwidth]{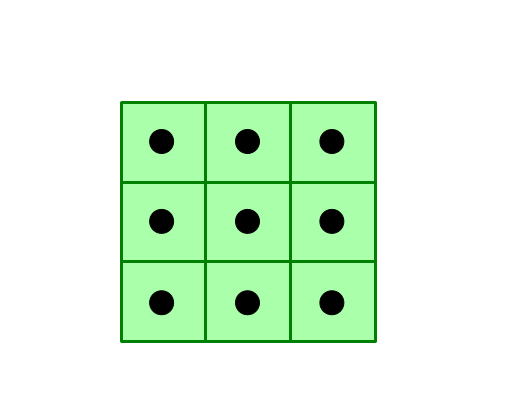}\label{fig: Fig. 4(a)}}%
		\subfloat[]{\includegraphics[width=0.14\textwidth]{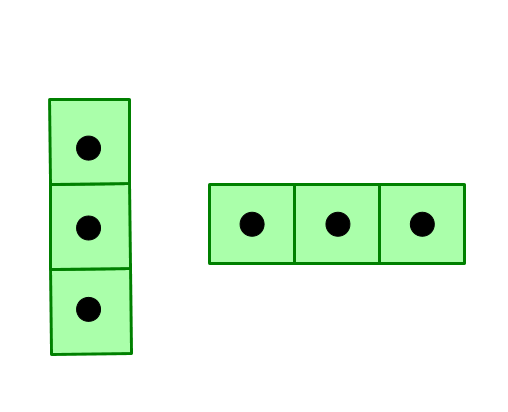}\label{fig: Fig. 4(b)}}\hfil
		\subfloat[]{\includegraphics[width=0.14\textwidth]{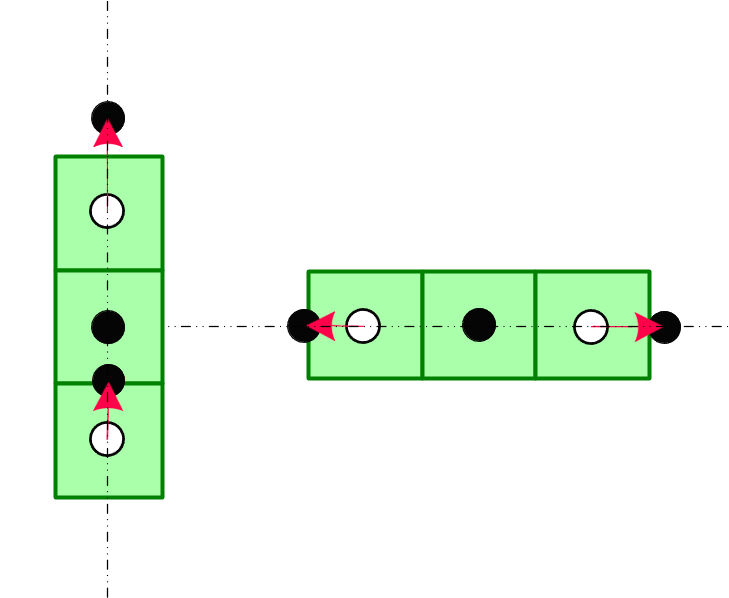}\label{fig: Fig. 4(c)}}
		\caption{(a) $3\times3$ regular convolution. (b) Factorized convolutions. (c) Factorized restricted deformable convolution. The nonlinearities in (b) and (c) are omitted in this illustration. The dark points, hollow circles and red arrows are the same as the definitions in Fig.~\ref{fig: Fig. 2}.}
		\label{fig: Fig. 4}
	\end{figure}
	2D filters can be approximated as a combination of 1D filters, for the sake of reducing memory and computational cost. In\cite{alvarez2016decomposeme}, a basic decomposed layer consists of vertical kernels followed by horizontal ones, and a nonlinearity is inserted in between 1D convolutions. For example, a convolutional layer of $3\times3$ (Fig.~\ref{fig: Fig. 4(a)}) can be decomposed into two consecutive factorized convolutional layers of $3\times1$ and $1\times3$ (Fig.~\ref{fig: Fig. 4(b)}). The ERFNet\cite{romera2017erfnet} has shown a good tradeoff between efficiency and accuracy with factorized convolutions. This paper also provides a factorized version of RDC.

	For 2D RDC, each learned offset has two components: vertical direction and horizontal direction. With 2D kernel decomposed into a vertical kernel and a horizontal kernel, the offsets can also be decomposed into two components of the same directions, as shown in Fig.~\ref{fig: Fig. 4(c)}. In 1D kernels, only one parameter is learned to control the dilation factor for each outer location. This is called Factorized Restricted Deformable Convolution (FRDC). It can also be interpreted as that the dilations in 1D kernels are adaptively learned. The number of additional parameters for FRDC is $\sqrt{N}-1$, only half of that of the RDC. That means FRDC is less flexible than RDC.
	
	\section{The Generation of Fisheye Image Dataset for Semantic Segmentation}
	Few large-scale datasets are available worldwide for fisheye image semantic segmentation. To enrich such datasets, a transformation method is proposed to convert conventional images to fisheye images. A mapping is built from the fisheye image plane to the conventional image plane. Thus the scene in conventional image can be remapped into fisheye image.
	\subsection{Mapping Conventional Image to Fisheye Image}
	A conventional image is captured from a pinhole camera. The perspective projection of a pinhole camera model can be described by (\ref{eqn:perspective}); for fisheye cameras, perhaps the most common model is the equidistance projection\cite{kannala2006generic}, as in (\ref{eqn:equidistance}).
	\begin{equation}
	\label{eqn:perspective}
	r=f\tan\theta
	\end{equation}
	\begin{equation}
	\label{eqn:equidistance}
	r=f\theta
	\end{equation}
	where $\theta$ is the angle between the principal axis and the incoming ray, $r$ is the distance between the image point and the principal point, and $f$ is the focal length. Both a conventional image and a fisheye image can be treated as a hemisphere image projected onto a plane according to different projection models and from different view angles. The details of the geometrical imaging model are described by\cite{miyamoto1964fish,kannala2006generic}. With the settings that the focal lengths of the perspective projection and the equidistance projection are identical and the max viewing angle $\theta_{max}$ is equal to $180^\circ$. The mapping from the fisheye image point $\mathbf{P}_f=(x_f,y_f)$ to the conventional image point $\mathbf{P}_c=(x_c,y_c)$ is described by
	\begin{equation}
	\label{eqn:mapping}
	r_c=f\tan(r_f/f)
	\end{equation}
	where $r_c=\sqrt{(x_c-u_{cx})^2+(y_c-u_{cy})^2}$ denotes the distance between the image point $\mathbf{P}_c$ and the principal point $\mathbf{U}_c=(u_{cx},u_{cy})$ in the conventional image, and $r_f=\sqrt{(x_f-u_{fx})^2+(y_f-u_{fy})^2}$ correspondingly denotes the distance between the image point $\mathbf{P}_f$ and the principal point $\mathbf{U}_f=(u_{fx},u_{fy})$ in the fisheye image.
	
	The mapping relationship (\ref{eqn:mapping}) is determined by the focal length $f$. A base focal length $f_0$ can be set. Thus the fisheye camera model approximately covers a hemispherical field. Each image and its corresponding annotation in the existing segmentation dataset are transformed using the same mapping function to generate the fisheye image dataset.
	
	The final mapping formula is formulated by
	\begin{equation}
	\label{eqn:final_mapping}
	\left\{
	\begin{array}{ccl}
	r_c=f\tan(\sqrt{(x_f-u_{fx})^2+(y_f-u_{fy})^2}/f)\\
	x_c=\text{sgn}(x_f)|r_c\cos(\text{atan2}(\frac{y_f - u_{fy}}{x_f - u_{fx}}))|+u_{cx} \\
	y_c=\text{sgn}(y_f)|r_c\sin(\text{atan2}(\frac{y_f - u_{fy}}{x_f - u_{fx}}))|+u_{cy}
	\end{array}
	\right.
	\end{equation}
	\subsection{Zoom Augmentation for Fisheye Image}
	Training of deep networks requires a huge number of training images, but training datasets are always limited. Data argumentation methods are adopted to enlarge training data using label-preserving transformations. Many forms are employed to do data augmentation for semantic segmentation, such as horizontally flipping, scaling, rotation, cropping and color jittering. Among them, scaling (zoom-in/zoom-out) is one of the most effective forms. DeepLab\cite{chen2016deeplab} augmented training data by random scaling the input images (from $0.5$ to $1.5$). PSPNet\cite{zhao2016pyramid} adopted random resize between $0.5$ and $2$ combining with other augmentation methods. Conventionally, scaling means the operation of changing the image's size. On the other hand, scaling the image can also be reasonably treated as the action of changing the focal length of the camera.
	
	Following this idea, is proposed in\cite{deng2017cnn} a new data augmentation method called zoom augmentation which is specially designed for fisheye images. Instead of simply resizing the image, the zoom augmentation means augmenting training dataset with additional data that is derived from an existing source by changing the focal length of the fisheye camera. Zoom augmentation adopts the mapping function in (\ref{eqn:mapping}).
	
\begin{figure*}[!t]
		\centering 
		\includegraphics[width=0.85\textwidth]{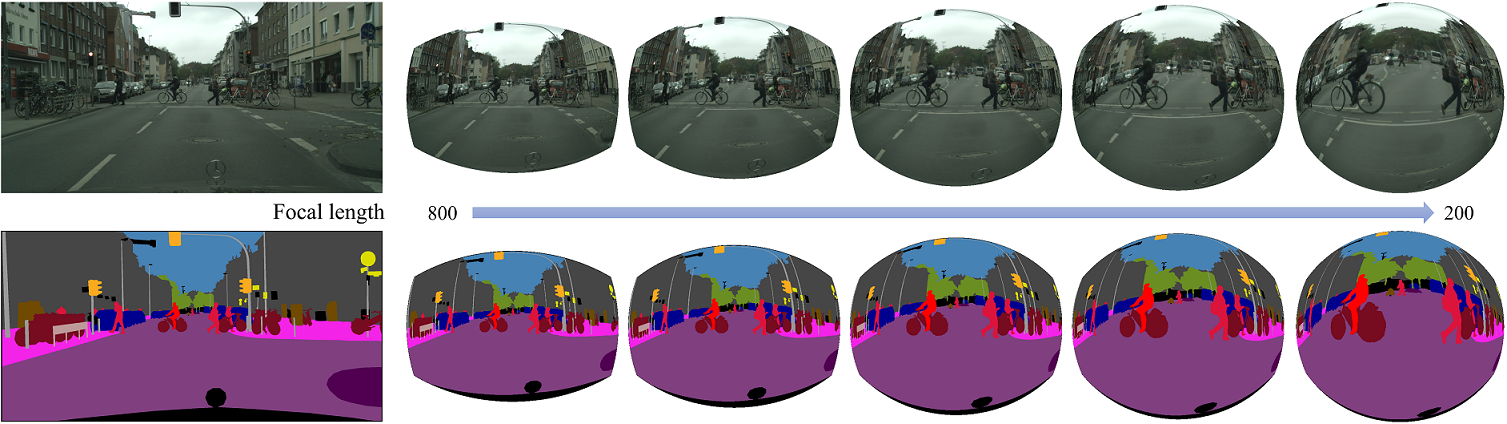}
		\caption{Zoom augmentation results. The left are the original color image and annotation. The right are the transformed images and annotations by zoom augmentation with a focal length changing from $200$ to $800$.}
		\label{fig: Fig. 5}
	\end{figure*}

	\cite{deng2017cnn} first transformed conventional images to fisheye images through a mapping function with a fixed focal length, then used zoom augmentation to augment the transformed images by changing the focal length. In this paper, the whole process is expressed in a united way. The operation of warping conventional images to fisheye-style images is generally called zoom augmentation. The zoom augmentation can adopt a fixed focal length or a randomly changing focal length. Via the zoom augmentation method, an existing conventional image dataset for semantic segmentation can be transformed into a fisheye-style image dataset. Fig.~\ref{fig: Fig. 5} illustrates how the focal length affects the mapping results. The smaller the focal length, the larger the degree of distortions. Thus, we can get fisheye images with different distortions by randomly changing the focal length.
	
	\begin{figure*}[!t]
		\centering
		\includegraphics[width=0.95\textwidth]{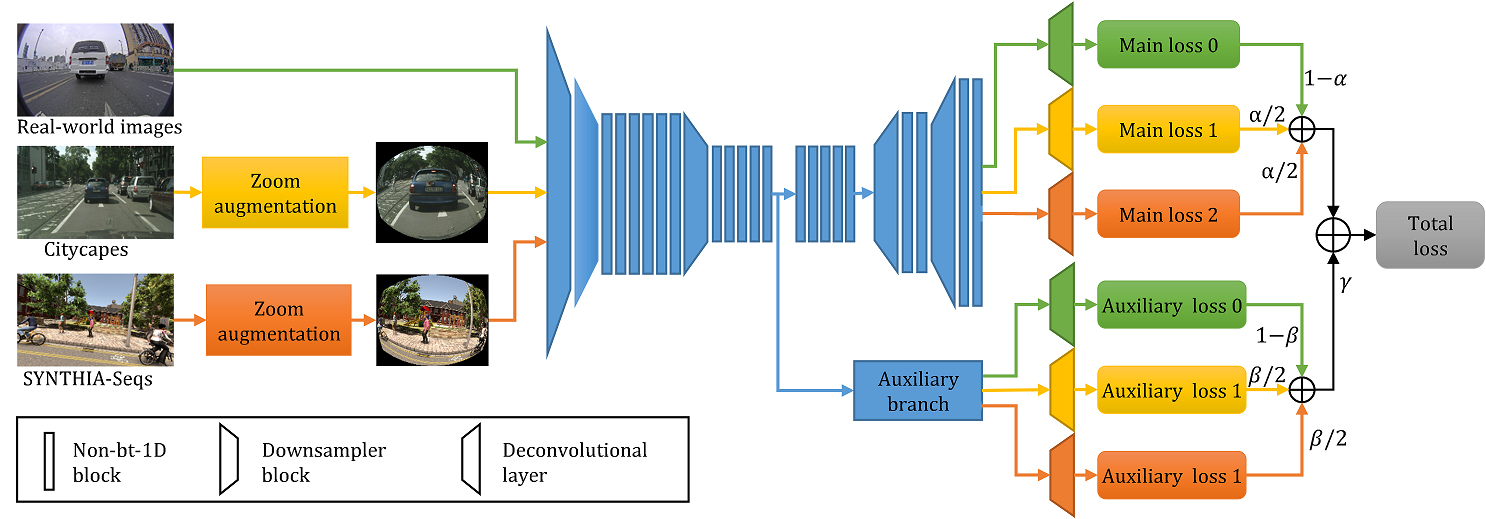}
		\caption{The multi-task learning architecture for road scene semantic segmentation. Cityscapes and SYNTIA-Seqs are transformed via zoom augmentation layers. The data are then fed into three shared-weight sub-networks (the blue blocks). The BN statistics are not shared among the sub-networks. The total loss is the weighted sum of main losses and auxiliary losses. $\gamma$ is auxiliary loss weighting to balance the contribution of auxiliary losses. $\alpha$ is the task weighting of main branch to balance the main losses of different tasks. Similarly, $\beta$ is the task weighting of auxiliary branch to balance the auxiliary losses of different task.}
		\label{fig: Fig. 6}
	\end{figure*}
	
	\section{Training Strategy}
	In this section, we introduce the strategy of training the CNN model to improve semantic segmentation accuracy on real-world surround view images with the help of transformed images. It is rarely practical to train the model using the transformed datasets and then use it to handle real-world images, due to different label spaces (Not all the target categories are the same as those of the source) and domain shift\cite{li2016revisiting} (different datasets). A simple way is to use a real-world dataset to fine-tune the CNN model pre-trained on transformed datasets. However, it risks overfitting when the amount of real-world images is limited. This paper uses both transformed images and real-world images to train the model. As shown in Fig.~\ref{fig: Fig. 6}, a multi-task learning architecture is built to train the model on datasets with different label spaces. The ERFNet is adopted as the base model. The last deconvolution layer in the ERFNet serves as a classifier. The same model with bounded weights is used to train both the source (transformed images) and the target (real-world images) domains. Two approaches are used to handle the domain shift and improve the generalization ability.
	\subsection{Sharing Weights with Private Batch Normalization Statistics}
	As illustrated in Fig.~\ref{fig: Fig. 6}, all the weights except those of classifiers are shared to learn domain-invariant features. Domain related knowledge is heavily related to the statistics of the Batch Normalization (BN) layer. In order to learn domain-invariant features, it is better for each domain to keep its own BN statistics in each layer. This paper uses an effective way for domain adaptation by sharing the weights but computing BN Statistics for each domain. This is called AdaBN in\cite{li2016revisiting}.
	\subsection{Hybrid Loss Weightings}
	During training, the loss function is the weighted sum of softmax losses of the three tasks, as in (\ref{eqn:mainloss}). $\alpha$ is the task weighting of the main branch to balance the main losses of different tasks. $\mloss{0}$ is the loss for real-world images. $\mloss{1}$ and $\mloss{2}$ are the losses for transformed images. The losses for transformed images act as a regularization term controlled by $\alpha$. A smaller $\alpha$ can make the training focus on the real data. A too-small $\alpha$ incurs model overfitting, whereas a too-large $\alpha$ has a consequence that the loss for real data will be overwhelmed by the regularization loss.
	
	Auxiliary loss is introduced into the net to help optimize the learning process, which is commonly used in the training process such as in PSPNet and GoogLeNet. The auxiliary loss enhances the backpropagation signal for the lower stages. We use the auxiliary losses to further balance the contribution of different losses. The auxiliary loss is formulated by (\ref{eqn:auxloss}). $\beta$ is the task weighting of the auxiliary branch to balance the auxiliary losses of different tasks. $K=2$ is the number of auxiliary tasks. $\aloss{0}$ is the auxiliary loss for real-world images. $\aloss{1}$ and $\aloss{2}$ are the auxiliary losses for transformed images. During training, the weighted auxiliary loss $\mathcal{L}_{aux}$ is added to the total loss with a discount weight $\gamma$, as formulated by (\ref{eqn:totalloss}). $\gamma$ is auxiliary loss weighting to balance the contribution of auxiliary loss.
	\begin{gather}
	\label{eqn:mainloss}
	\mathcal{L}_{main}=(1.0-\alpha)\mloss{0}+\frac{\alpha}{K}\sum_{i=1}^K\mloss{i}\\
	\label{eqn:auxloss}
	\mathcal{L}_{aux}=(1.0-\beta)\aloss{0}+\frac{\beta}{K}\sum_{i=1}^K\aloss{i}\\
	\label{eqn:totalloss}
	\mathcal{L}_{total}=\mathcal{L}_{main}+\gamma\mathcal{L}_{aux}
	\end{gather}
	In this paper, we use different task weightings for the main branch loss $\mathcal{L}_{main}$ and the auxiliary branch loss $\mathcal{L}_{aux}$. That means, $\alpha$ does not have to equal to $\beta$. A bigger weighting $\beta$ can introduce stronger regularization. Thus with the bigger $\beta$ for auxiliary loss, a smaller $\alpha$ for main loss can be employed to balance the contribution of different losses. This method is termed as Hybrid Loss Weightings (HLW).
	\section{Experiments}
	In this section, the datasets used for the experiments are first introduced. Then the RDC based model is evaluated on conventional and transformed fisheye datasets, respectively. In the end, experiments for road scene semantic segmentation using surround view cameras are conducted. A platform with two NVIDIA GTX 1080Ti GPUs is used to train and evaluate the models using MXNet. The semantic segmentation performance is measured by the standard metric of mean Intersection-Over-Union (mIoU).
	
	\subsection{Datasets for Surround View Image Semantic Segmentation}
\begin{figure}[!t]
		\centering
		\begin{minipage}[tb]{0.9\columnwidth}
			\includegraphics[width=0.23\columnwidth]{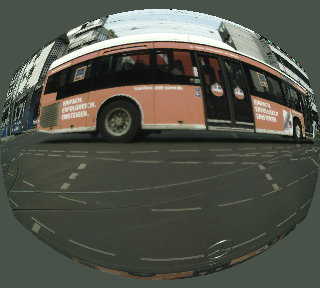}
			\includegraphics[width=0.23\columnwidth]{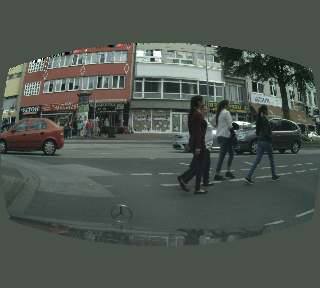}
			\includegraphics[width=0.23\columnwidth]{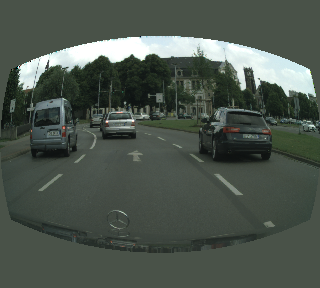}
			\includegraphics[width=0.23\columnwidth]{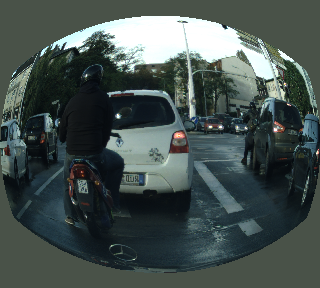}\vspace{2pt}
			
			\includegraphics[width=0.23\columnwidth]{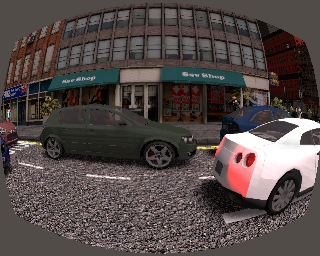}
			\includegraphics[width=0.23\columnwidth]{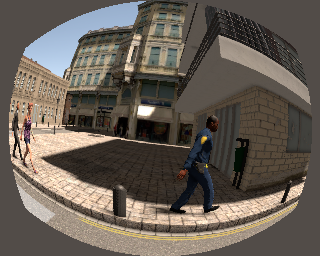}
			\includegraphics[width=0.23\columnwidth]{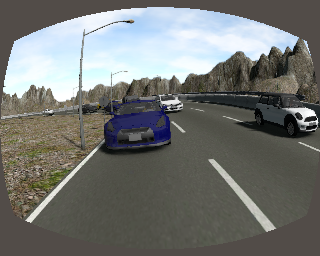}
			\includegraphics[width=0.23\columnwidth]{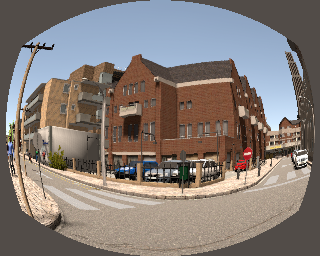}
		\end{minipage}
		\caption{Transformed images by zoom augmentation with a randomly changing focal length. Images in the first row are from Cityscapes. Images in the second row are from SYNTHIA-Seqs. The first row captures the scene of the front view. The second row captures scene with different perspectives.}
		\label{fig: Fig. 7}
	\end{figure}

	Two complementary datasets Cityscapes\cite{cordts2016cityscapes} and SYNTHIA-Seqs\cite{ros2016synthia} are used to augment surround view datasets via the zoom augmentation method. Cityscapes is a real large-scale dataset captured by a forward-looking conventional camera. $3475$ images are used in the experiments. SYNTHIA-Seqs are captured in a virtual city using four conventional cameras of different directions. Specifically, the sub-sequences: Spring, Summer, and Fall of SEQS-01, SEQS-02 and SEQS-04 are used, totally containing 34696 images. Both the real and synthetic datasets are transformed to augment surround view images. The resolutions of the transformed images after applying zoom augmentation are set to $576\times640$ and $512\times640$ for Cityscapes and SYNTHIA-Seqs, respectively. In order to employ zoom augmentation with a randomly changing focal length for the training, this paper implements a zoom augmentation layer. This new layer adopts a CUDA implementation. Thus the image can be transformed online. The time consumption of the layer is very small. Some examples transformed by the zoom augmentation method are illustrated in Fig.~\ref{fig: Fig. 7}.
	
	Besides, $600$ surround view images are captured by four fisheye cameras mounted around a moving vehicle and annotated using the Cityscapes annotation tool\cite{cordts2016cityscapes}. Of these, $350$ images are used for training, $100$ images are used for validation, and $150$ images are used for testing. The defined classes are listed in Fig.~\ref{exp_real_compare}(b). Each image and annotation have a resolution of $512\times864$. This dataset is denoted as SVScape dataset. Some examples are shown in Fig.~\ref{exp_real_compare}.
	
	\subsection{Evaluation for the Restricted Deformable Convolution}
	The cityscapes dataset and a Fisheye-Cityscapes are used to analyze the performances of different models on conventional images and fisheye images. The Fisheye-Cityscapes dataset is generated from the Cityscapes dataset generated by the zoom augmentation method with a fixed focal length of $159$. The images of Cityscapes dataset are resized to $512\times1024$. The resolution of Fisheye-Cityscapes is $576\times640$. The datasets are respectively split into three sets for the ablation study: 2475 for training, 500 for validation, and 500 for testing.
	
\begin{figure}[!t]
		\centering
		\subfloat[]{\includegraphics[width=0.15\columnwidth,origin=c]{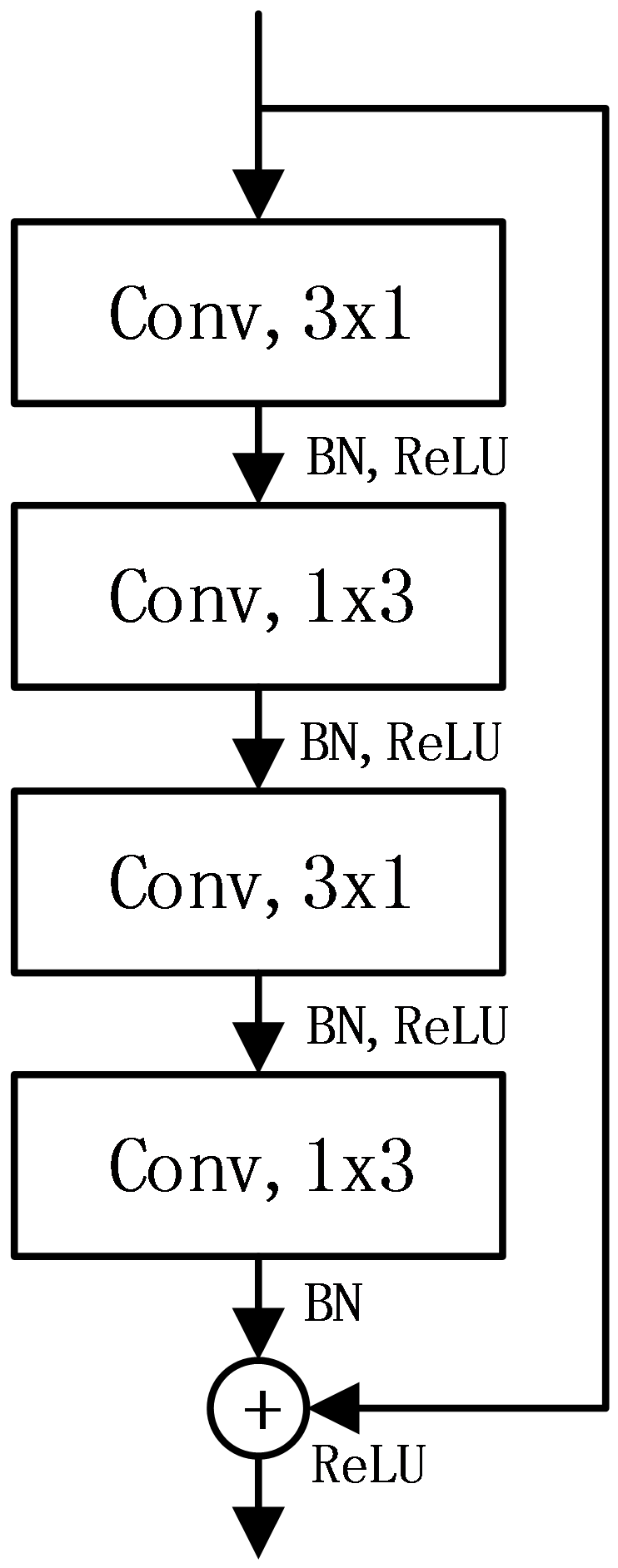}\label{fig: Fig. 8(a)}}\hfill
		\subfloat[]{\includegraphics[width=0.15\columnwidth,origin=c]{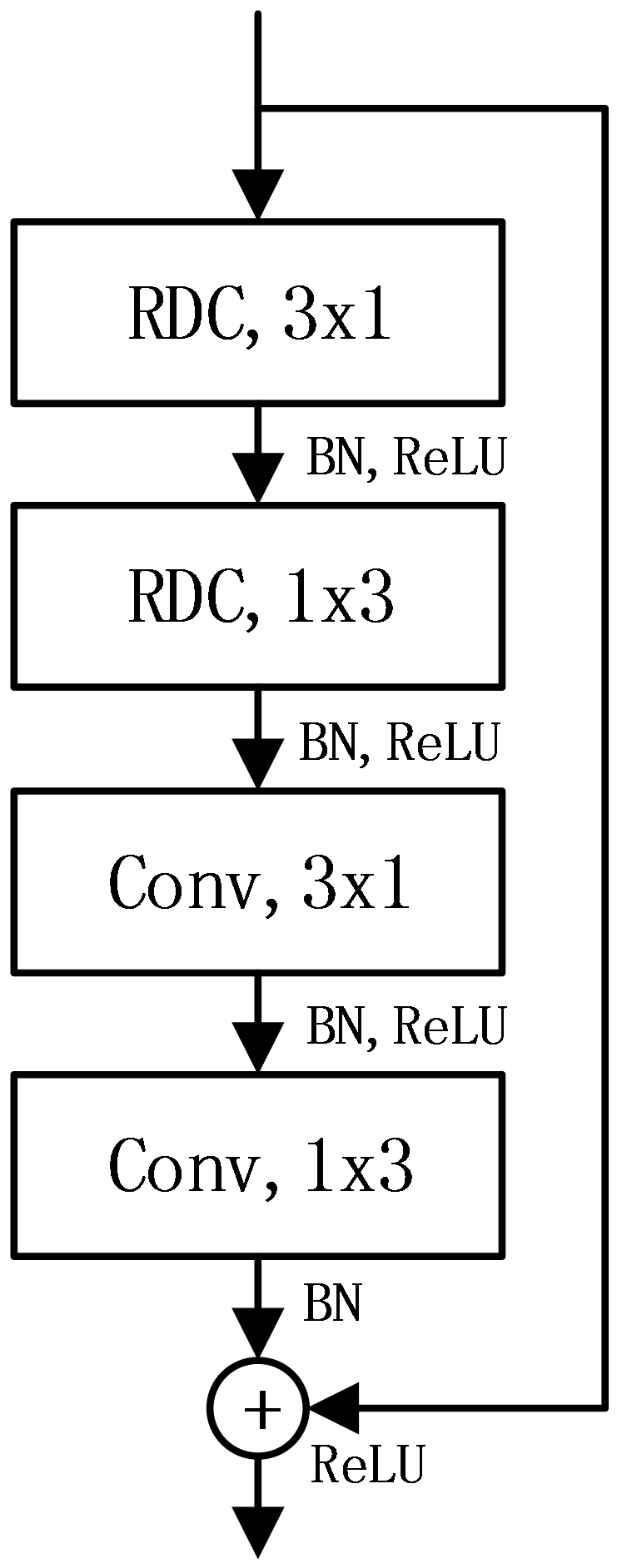}\label{fig: Fig. 8(b)}}\hfill
		\subfloat[]{\includegraphics[width=0.67\columnwidth,origin=c]{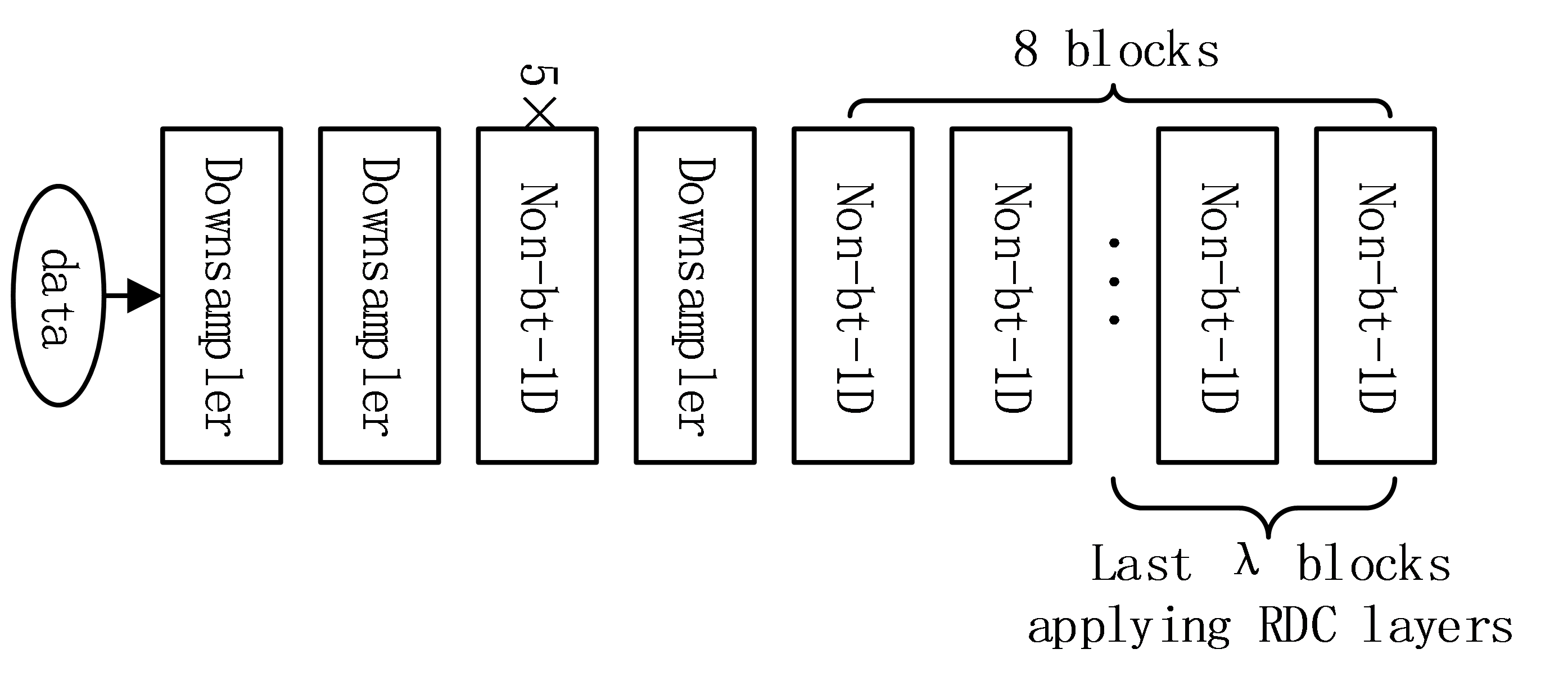}\label{fig: Fig. 8(c)}}
		\caption{The illustrate of ERFNet-RDC-$\lambda$. (a) Non-bt-1D block in ERFNet. (b) Reconstructed non-bt-1D block. The first two convolutional layers are replaced with RDC layers. (c) The encoder of ERFNet-RDC-$\lambda$.}
		\label{fig: Fig. 8}
	\end{figure}

	We evaluate the performance of restricted deformable convolution on VGG-based architecture and residual block-based architecture. The FCN-VGG16, ERFNet, ENet, ERFNet-OPP, and DeepLab are used for the evaluation. ERFNet-OPP denotes the ERFNet with OPP module attached between the encoder and decoder. A $1\times1$ convolutional layer is following the OPP module to reduce the dimension of the feature map. For clarity, let $\mathcal{B}$-$\mathcal{C}$-$\lambda$ denotes the modified models. $\mathcal{B}$ denotes the base model, including FCN-VGG16, ERFNet, ENet, and DeepLab. $\mathcal{C}$ denotes the type of modified convolutional layer, including DC (deformable convolution), RDC (restricted deformable convolution) and FRDC (factorized restricted deformable convolution). $\lambda$ is the numbers of reconstructed blocks or reconstructed stages. The last $\lambda$ blocks or stages are replaced by the reconstructed ones. The ERFNet is reimplemented in MXNet as the baseline model with a few differences. Batch normalization layer is applied after each convolutional layer and all the deconvolution layers use a kernel of $2\times2$ and stride $2$. The reconstructed block is built by replacing the first two convolutional layers in non-bt-1D block with modified convolutional layers. Fig.~\ref{fig: Fig. 8} illustrates the ERFNet-RDC-$\lambda$. The last $\lambda$ blocks in the encoder are replaced by reconstructed blocks. Similarly, for ENet, the reconstructed block is built by replacing the non $1\times1$ kernel convolutional layers in its third stage's ``bottleneck2.x" modules with modified layers. In FCN-VGG16, the convolutional layers of the frontend VGG-16 can be divided into five stages by ``maxpool" (Please refer to the Table 1 of \cite{simonyan2014very}). A reconstructed stage is built by replacing the $3\times3$ convolutional layers in a stage with DC or RDC layers. In FCN-VGG16-DC-$\lambda$ and FCN-VGG16-RDC-$\lambda$, the last $\lambda$ stages are replaced by the reconstructed stages.
	
	ERFNet, ENet, ERFNet-OPP, and their modified models are trained from scratch in MxNet using Nesterov Accelerated Gradient (NAG) with a mini-batch of $12$, momentum of $0.9$ and weight decay of $0.0002$. The initial learning rate is set to $0.05$. Class balancing is not applied in the experiments. Instead, the softmax loss is multiplied by $2.0$ to balance the regularization. Following the practice suggested in\cite{romera2017erfnet}, we first train the encoder and then attach the decoder to jointly train the full network. The “poly” learning rate policy (the learning rate is multiplied by $(1-\frac{iter}{max\_iter})^{power}$) is adopted to speed up the training. The power is set to $0.9$. The encoder is trained for $120$ epochs, and the joint model is trained for $100$ epochs. For the modified models, during training of the encoder, weights of the convolutional layer for offset learning are initialized to zero, yet other layers are initialized by the Xavier method. Unlike the training in\cite{dai2017deformable}, the encoder is not initiated with a pre-rained model. In order to stabilize the training, the offsets are kept unchanged in the first $20$ epochs. Thus a fixed conv shape is employed to warm up the training. Besides, the learning rates for offset learning are set to $1.0$ and $0.1$ times the base learning rate to train the encoder and joint model, respectively. 
	
	The FCN-VGG16 and its modified models are trained using the all-at-once strategy by scaling the skip connections with a fixed constant. The net is trained using SGD with a mini-batch size of 2, a learning rate of $0.0001$, a momentum of $0.98$, and a weight decay of $0.0005$. The weights of the Imagenet-pretrained VGG-16 is used to initialize the models. Training proceeds for $100$ epochs total.

\begin{figure}[!tb]
		\centering
		\includegraphics[width=0.50\columnwidth]{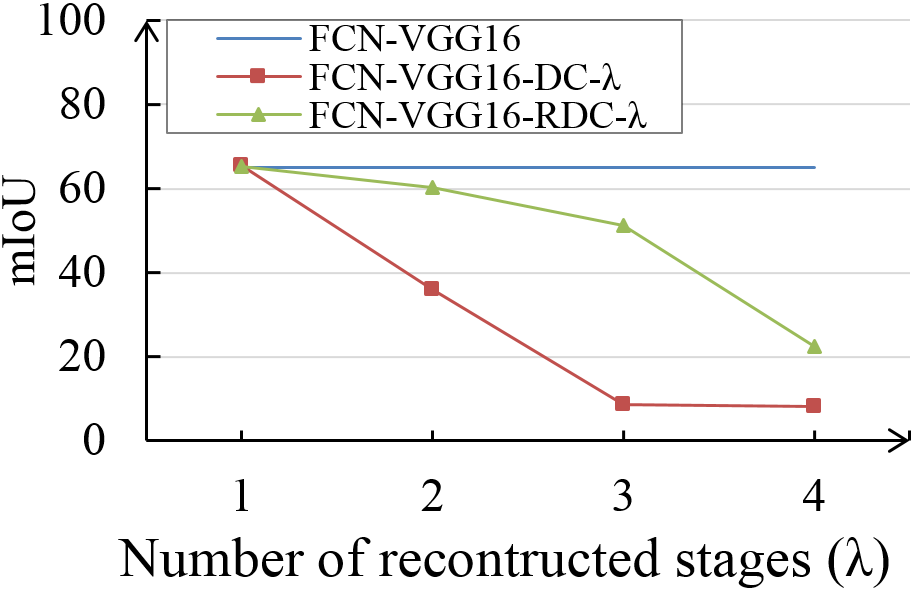}
		\caption{Results of VGG based architecture on test split of Cityscapes dataset. The accuracies largely decrease and even collapse when more DC or RDC layers are applied in FCN-VGG16. RDC alleviates the degradation by fixing the central sampling point.}
		\label{exp_fcn}
	\end{figure}

	\textbf{Performance evaluation on VGG-based architecture:} The results of FCN-VGG16 and its variants on the test split of Cityscapes dataset are shown in Fig.~\ref{exp_fcn}. FCN-VGG16-DC and FCN-VGG16-RDC based models show slight improvements when $\lambda$=$1$. However, the accuracies decrease largely and even collapse when $\lambda$ increases. The accuracy of FCN-VGG16-RDC decreases much slower than that of FCN-VGG16-DC. It shows that the performance degrades when DC or RDC is applied in the finer layers due to the spatial correspondence problem. RDC alleviates this problem by fixing the central sampling point but still get a bad performance.
	
	\begin{figure}[!t]
		\centering
		\subfloat[ERFNet]{\includegraphics[width=0.495\columnwidth]{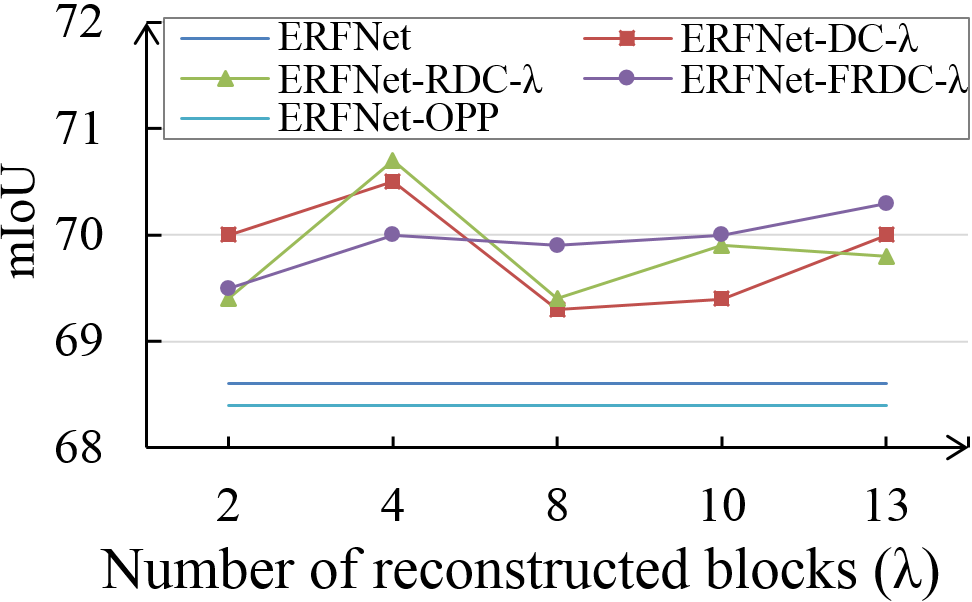}
			\label{exp_residual_cityscapes(a)}}
		\subfloat[ENet]{\includegraphics[width=0.495\columnwidth]{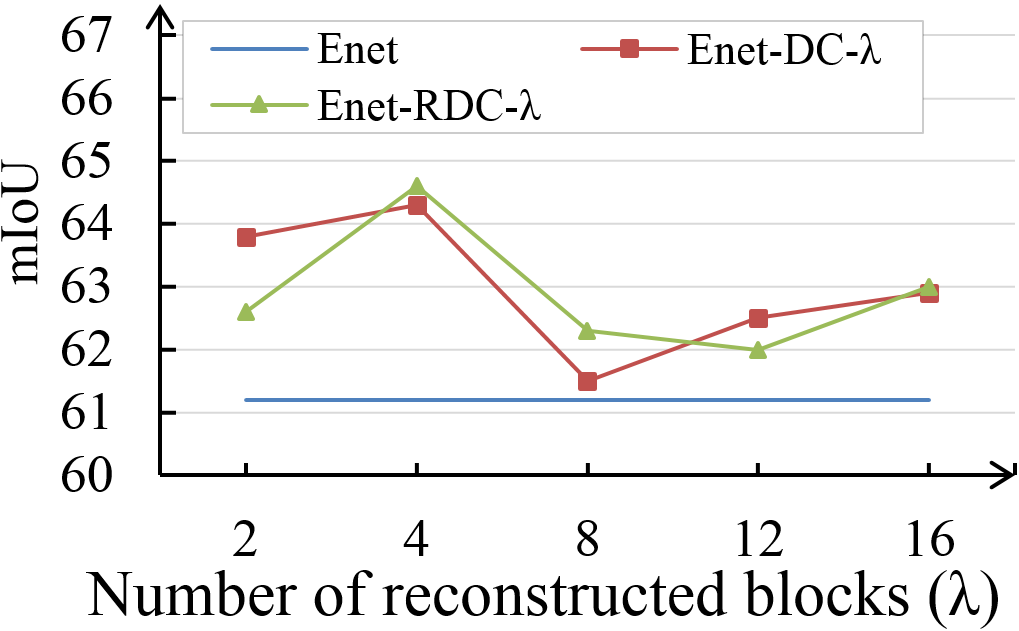}
			\label{exp_residual_cityscapes(b)}}
		\caption{Results of residual block-based architecture on the test split of Cityscapes dataset. The DC, RDC, and FRDC based models outperform the base models. ERFNet-RDC-4 and ENet-RDC-4 achieved the best accuracies, respectively.}
		\label{exp_residual_cityscapes}
	\end{figure}
	\begin{figure}[!t]
		\centering
		\subfloat[ERFNet]{\includegraphics[width=0.495\columnwidth]{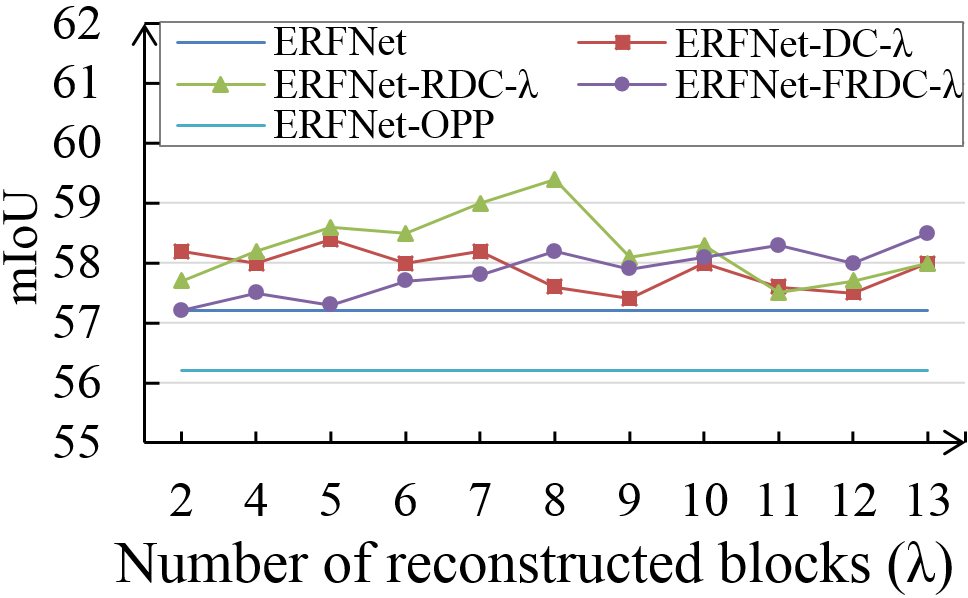}
			\label{exp_residual_fisheye(a)}}
		\subfloat[ENet]{\includegraphics[width=0.495\columnwidth]{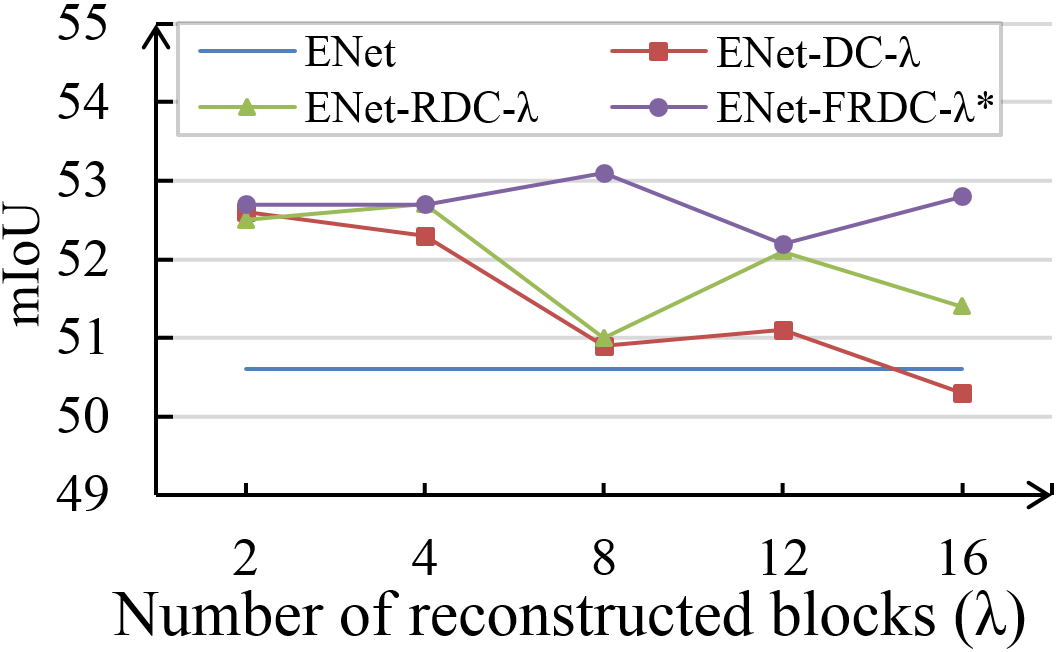}
			\label{exp_residual_fisheye(b)}}
		\caption{Results of residual block-based architecture on the test split of Fisheye-Cityscapes dataset. The DC, RDC, and FRDC based models almost outperform the base models. As with the increase of $\lambda$, the accuracies of the models first improve and then become saturated. The ERFNet-RDC-8 and ENet-FRDC-8* achieved the best accuracies, respectively.}
		\label{exp_residual_fisheye}
	\end{figure}
	
	\textbf{Performance evaluation on residual block-based architecture:} The results of ERFNet, Enet and their variants on the test sets of Cityscapes and Fisheye-Cityscapes are shown in Fig.~\ref{exp_residual_cityscapes} and Fig.~\ref{exp_residual_fisheye}. The ERFNet-OPP produces worse accuracies than ERFNet as shown in Fig.~\ref{exp_residual_cityscapes}(a) and Fig.~\ref{exp_residual_fisheye}(a). The ERFNet is a highly compact and efficient network constructed with residual blocks. We conjecture that the OPP module\cite{deng2017cnn} may increase learning difficulty or impact gradient propagation from decoder to encoder. Similarly, the experiments of S\'{a}ez et al.\cite{saez2018cnn} shows the pyramid pooling module of PSPNet also did not improve ERFNet. It suggests a more effective module is required to improve ERFNet.
	Fig.~\ref{exp_residual_cityscapes} and Fig.~\ref{exp_residual_fisheye} show the DC, RDC, and FRDC based models are almost better than the base models. Considering the results of VGG-based architecture, it suggests that the identity mapping of residual block is critical to prevent the modified models from collapsing. However, as with the increase of $\lambda$, the performances of these models first improve and then become saturated.
	
\begin{figure}[!tb]
		\centering
		\includegraphics[width=0.7\columnwidth]{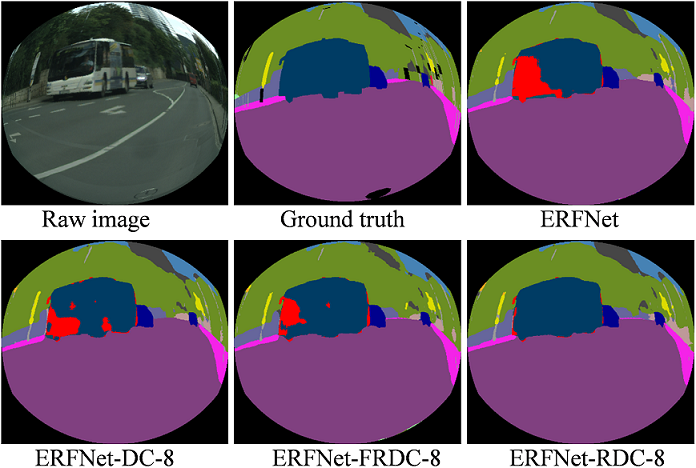}
		\caption{One example of the segmented results produced by ERFNet, ERFNet-DC-8, ERFNet-FRDC-8, ERFNet-RDC-8. The red pixels denotes false recognitions of the bus. The ERFNet-RDC-8 nearly detected the whole bus in the image.}
		\label{fisheye_compare}
	\end{figure}

	When $\lambda$=$2$, DC based models achieve better accuracies in both conventional dataset (Fig.~\ref{exp_residual_cityscapes}) and fisheye-style dataset (Fig.~\ref{exp_residual_fisheye}) than other models. As with the increase of $\lambda$, RDC based models outperform the DC based models. However, both of them show degradation when the $\lambda$ increases. FRDC based models constantly show small improvements as shown in Fig.~\ref{exp_residual_cityscapes}(a) and Fig.~\ref{exp_residual_fisheye}(a). 
	The ENet-FRDC* in Fig.~\ref{exp_residual_fisheye}(b) is built upon the ENet-RDC instead of ENet. The FRDC is defined on the factorized convolutions. So, we apply the FRDC into the factorized convolutional layers of ENet-RDC where RDC layers are applied. Thus, we can observe how the FRDC impacts performance. Note that the ENet adopts a large kernel size, $5\times1$ and $1\times5$, for factorized filters, which needs more parameters learned to define the kernel shape for RDC layers. As shown in Fig.~\ref{exp_residual_fisheye}(b), the accuracy of ENet-FRDC* has higher accuracy than ENet-RDC when $\lambda$ increases. The experimental results show that restricting the deformable convolution is effective for the semantic segmentation. The DC and RDC essentially possess a more powerful ability to model geometric transformations because there are no constraints for the outer sampling locations of RDC and DC kernels. The RDC based model is less prone to saturation than the DC based model and achieves better accuracy. As shown in Fig.~\ref{exp_residual_fisheye}(a), ERFNet-RDC-8 achieved the best score on the Fisheye-Cityscapes dataset, which is adopted in the next section. Fig.~\ref{fisheye_compare} shows an example of the results by different models.
	
	\textbf{Performance evaluation on deeper residual block-based architecture:} We evaluate the performance on DeepLab, which uses ResNet-101 feature extraction network. DeepLab-DC-$\lambda$ and DeepLab-RDC-$\lambda$ are used to denote the base DeepLab model with DC and RDC layers. $\lambda$ means that last $\lambda$ $3\times3$ convolutional layers are replaced by DC or RDC layers. The training process and hyper-parameter settings are following \cite{dai2017deformable}. ImageNet-pretrained ResNet-101 is used to initialize the weights. Two GPUs are used to train the models on the original Cityscapes dataset the resolution of which is $1024\times2048$. The models are trained for 106 epochs with a batch size of 4 on randomly sampled crops of size $768\times1024$. The base model employs atrous convolution with dilation $2$ in the last three $3\times3$ convolutional layers. As shown in Table~\ref{tab_deeplab}, when enlarging the dilation ratio, accuracy increases. Although large dilation ratios improve the base model, performance gains more when using DC or RDC layers with dilation $2$. When $\lambda$ is greater than or equal to $3$, the model employed RDC layers is constantly better than that employed DC layers. When $\lambda=7$, DeepLab-RDC achieves the best performance. We note that when $\lambda=9$, the DeepLab-DC shows large performance degradation, while the DeepLab-RDC still shows acceptable performance. We infer that the model with DC layers is less stable than that with RDC layers during training when more deformable filters are employed. For more details about the properties of deformable filters, we refer readers to Dai et al.\cite{dai2017deformable}.
	
	\begin{table}[tb]
		\renewcommand{\arraystretch}{1.3}
		\centering
		\caption{Results of employing DC or RDC in the last two stages of ResNet-101 feature extraction network, and employing different dilation ratios in the base model on the validation/test set of original Cityscapes dataset.}
		\label{tab_deeplab}
		\begin{tabular}{c|c|c} 
			\hline
			\tabincell{c}{$\lambda$ and dilation ratios of \\last three $3\times3$ convolutions}
			& \tabincell{c}{DeepLab-DC-$\lambda$ \\ mIoU@val/test}
			& \tabincell{c}{DeepLab-RDC-$\lambda$\\mIoU@val/test}\\
			\hline\hline
			$\lambda=0$\footnotemark[1], dilation(2,2,2) & \multicolumn{2}{c}{67.9/66.7} \\
			\hline
			$\lambda=0$, dilation(4,4,4) & \multicolumn{2}{c}{69.4/-} \\
			\hline
			$\lambda=0$, dilation(6,6,6) & \multicolumn{2}{c}{$\mathbf{70.8/69.5}$} \\
			\hline
			$\lambda=0$, dilation(8,8,8) & \multicolumn{2}{c}{70.4/-} \\
			\hline
			$\lambda=1$, dilation(2,2,2) & 71.9/- & 71.1/- \\
			\hline
			$\lambda=3$, dilation(2,2,2) & $\mathbf{73.7/71.6}$ & 73.8/- \\
			\hline
			$\lambda=5$, dilation(2,2,2) & 72.5/- & 72.8/- \\
			\hline
			$\lambda=7$, dilation(2,2,2) & 73.4/- & $\mathbf{74.0/72.0}$ \\
			\hline
			$\lambda=9$, dilation(2,2,2) & 58.6/- & 73.8/- \\
			\hline
			$\lambda=11$, dilation(2,2,2) & 68.1/- & 73.9/- \\
			\hline
			$\lambda=20$, dilation(2,2,2) & 70.1/- & 74.0/- \\
			\hline
			$\lambda=26$, dilation(2,2,2) & 73.0/- & 73.2/- \\
			\hline
		\end{tabular}\\
		\footnotemark[1]{$\lambda=0$ means the base DeepLab model without DC or RDC layers.}
	\end{table}
	
	\subsection{Semantic Segmentation Using Surround View Cameras}
	The multi-task learning architecture is shown in Fig.~\ref{fig: Fig. 6}. The ERFNet-RDC-8 is adopted as the base model. The auxiliary branch is a convolutional layer with a kernel of $1\times1$, stride 1, and $128$ output channels. Batch Normalization and ReLU are applied after this layer. Main loss weighting $\alpha$ and auxiliary loss weighting $\beta$ are set to balance the contribution of real samples and transformed samples. The net is trained with the training set of SVScape, Cityscapes, and SYNTHIA-Seqs. The weights except those of the classifiers are shared among all the tasks. The training procedure follows the ERFNet-RDC described in the previous section. 50K iterations are employed for training encoder and the joint model, respectively. In each iteration, four images are drawn from the three datasets to generate a mini-batch of $12$ samples. The conventional images of Cityscapes and SYNTHIA-Seqs are transformed to fisheye images online through the zoom augmentation layer. The zoom augmentation method can adopt a fixed focal length or a randomly changing focal length. For the fixed mode, the focal length is set to $240$ and $300$ for Cityscapes and SYNTHIA-Seqs. For random mode, the focal length is changed randomly between $200$ and $800$.

	After introducing an auxiliary branch, a strategy of HLW that employs different loss weightings is applied. That means main loss weighting $\alpha$ does not have to be equal to auxiliary loss weighting $\beta$. $\gamma$, $\alpha$, $\beta$ are determined by experimenting with different weighting values on the validation dataset of SVScape. AdaBN and zoom augmentation with a randomly changing focal length are employed in the experiments. We first determine the value of $\gamma$ by fixing the $\alpha$ and $\beta$ to $\frac{1}{2}$. Then determine the $\alpha$ and $\beta$ with the selected $\gamma$. Table~\ref{tab_weighting} shows the results. The accuracy is improved with a broad value of $\gamma$. $\gamma=0.3$ yields the best performance when $\alpha=\beta=\frac{1}{2}$. 
	The $\gamma$ is fixed to $0.3$ for the subsequent testing. When decreasing $\alpha$ with $\alpha$ equal to $\beta$, the model shows performance degradation, because the generalization ability of the model becomes poor due to weak regularization from auxiliary tasks. When setting a smaller $\alpha$ and a bigger $\beta$, the performance shows a significant improvement. When $\alpha=\frac{1}{3}$ and $\beta=\frac{1}{2}$, the model achieved the best performance. Thus, $\gamma=0.3$, $\alpha=\frac{1}{3}$, $\beta=\frac{1}{2}$ are set for the strategy of HLW.
	
	\begin{table}[!t]
		\centering
		\renewcommand{\arraystretch}{1.3}
		\caption{Loss Weightings Determination on Validation Set of SVScape}
		\label{tab_weighting}
		\begin{tabu} to 0.8\columnwidth{X[1,c]|X[1,c]|X[1,c]|X[1,c]}
			\hline
			$\alpha$ & $\beta$ & $\gamma$ & mIoU ($\%$)\\
			\hline
			\hline
			\ts{1/2}{0}{0}{70.3}\footnotemark[1] \\
			\ts{1/2}{1/2}{0.1}{71.4} \\
			\ts{1/2}{1/2}{0.2}{71.9} \\
			\ts{1/2}{1/2}{\mathbf{0.3}}{\mathbf{72.2}}\\
			\ts{1/2}{1/2}{0.4}{72.1}\\
			\ts{1/2}{1/2}{0.5}{71.2}\\
			\hline
			\ts{1/3}{1/3}{0.3}{72.3}\\
			\ts{\mathbf{1/3}}{\mathbf{1/2}}{0.3}{\mathbf{73.2}}\\
			\ts{1/3}{2/3}{0.3}{72.7}\\
			\hline
			\ts{1/5}{1/5}{0.3}{71.5}\\
			\ts{1/5}{1/2}{0.3}{72.4}\\
			\ts{1/5}{2/3}{0.3}{\mathbf{73.1}}\\
			\ts{1/5}{4/5}{0.3}{72.9}\\
			\hline
		\end{tabu}\\
		\footnotemark[1]{$\beta=0$ and $\gamma=0$ means plain multi-task learning with no auxiliary branch.}
	\end{table}
	
	\begin{table}[!t]
		\centering
		\renewcommand{\arraystretch}{1.3}
		\caption{Evaluation on the Test Set of SVScape using ERFNet-RDC-8}
		\label{tab_test}
		\begin{tabu} to 0.8\columnwidth{X[c]|X[3,c]|X[1,c]|X[1.5,c]}
			\hline
			AdaBn & Zoom augmentation & HLW & mIoU (\%)\\
			\hline
			\hline
			$\times$&	Fixed	& $\times$	& $64.6$ \\
			$\surd$&	Fixed	& $\times$	& $72.1$ \\
			$\surd$&	Random	& $\times$	& $72.6$ \\
			$\surd$&    $\times$& $\times$  & 71.7\\
			$\surd$&	Random	& $\surd$	& $\mathbf{74.2}$\\
			\hline
		\end{tabu}
	\end{table}	
	
	\begin{table}[!t]
		\renewcommand{\arraystretch}{1.3}
		\caption{Forward Pass Time for A $512\times864$ Image}
		\label{tab: time}
		\centering
		\begin{tabular}{c|c}
			\hline
			Model & net.forward (s)\\
			\hline
			\hline
			ERFNet&$0.016$\\
			ERFNet-RDC-8&$0.018$\\
			\hline
		\end{tabular}
	\end{table}

	We evaluated how AdaBN, zoom augmentation, and HLW affects performance on the test set of SVScape. The experimental results are reported in Table~\ref{tab_test}. When applied the AdaBN, the BN statistics are not shared and each BN layer computed BN statistics for each domain. Adopting the AdaBN largely improves the performance by $7.5\%$. The shape is important for the understanding of semantics, which is robust to the domain shift as argued in \cite{saleh2018effective}. Saleh et al.\cite{saleh2018effective} successfully leveraged shape cues to bridge the gap between the synthetic and real domains for better foreground-class segmentation. We use the zoom augmentation method to reduce shape differences between conventional images and fisheye images. The zoom augmentation changes the shapes of the semantics in conventional images to imitate those in fisheye images, as demonstrated in Fig.~\ref{fig: Fig. 5} and Fig.~\ref{fig: Fig. 7}. As shown in Table~\ref{tab_test}, zoom augmentation with randomly changing focal length brings $0.5\%$ improvement over the fixed focal length, which indicates training images with different degrees of distortion helps to improve the generalization ability. If disabling the zoom augmentation, which means that the conventional images go straight into the net, the accuracy drops by $0.9\%$. When HLW is not employed, $\alpha$ is set to $\frac{1}{3}$, and $\beta$, $\gamma$ are set to zero (Plain multi-task learning with no auxiliary branch). After employing HLW strategy with $\alpha=\frac{1}{3}$, $\beta=\frac{1}{2}$, and $\gamma=0.3$, the performance improves to $74.2\%$.
	
\begin{figure*}[htbp]
		\centering
		\subfloat[Results from different models]{\includegraphics[width=0.8\textwidth]{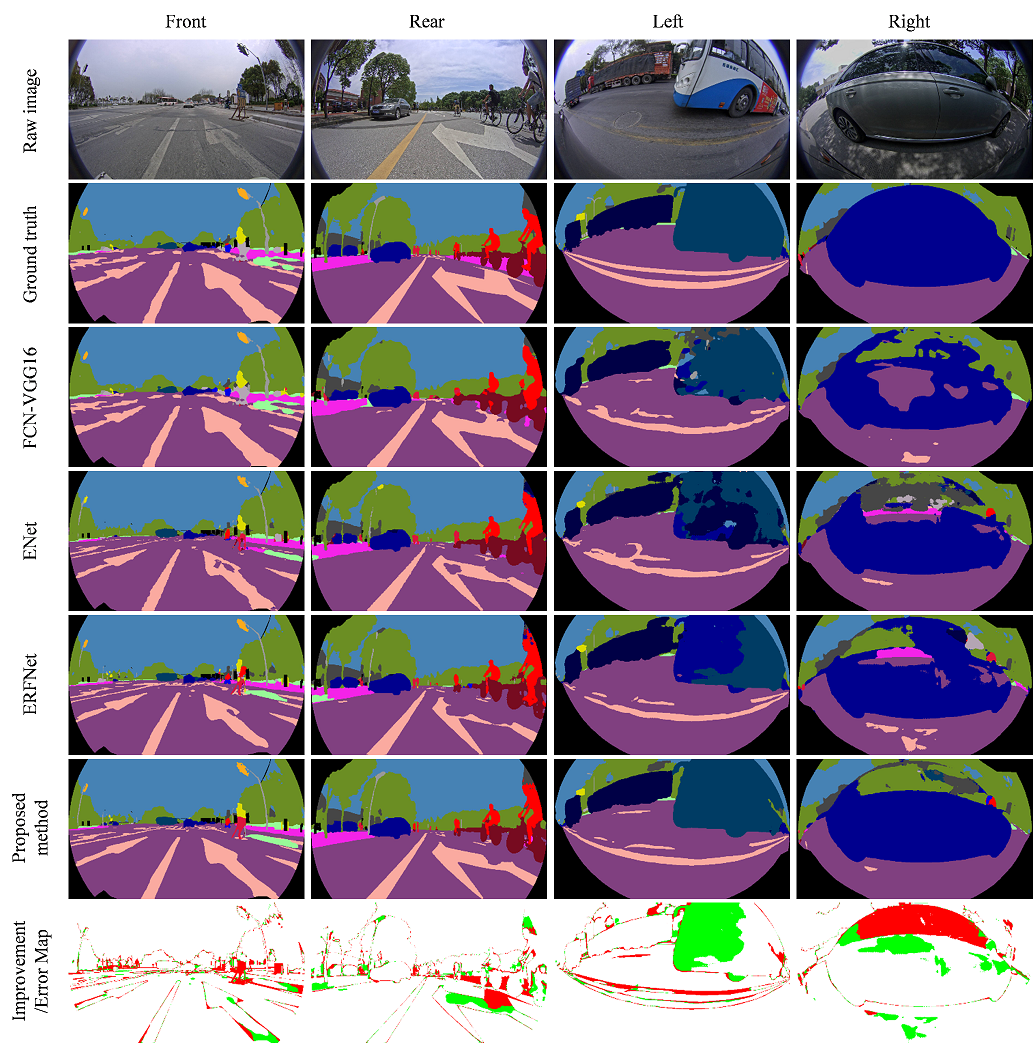}}\\
		\subfloat[List of 18 classes names and their corresponding colors used for labeling. The void class is marked as black. The vertical separator means a standing structure that used to separate areas, such as walls, fences and guard rails. Road markings which painted on the road surface are used to convey traffic information, usually including white or yellow lines or patterns. We believe this class is beneficial to solutions that use surround view cameras. Other classes adopt the same definitions as those in Cityscapes. Other unclear or ignored objects are assigned a void label, e.g., the reverse sides of traffic signs, commercial signs, electric wires and the invalid boundaries of the images.]{\includegraphics[width=0.95\textwidth]{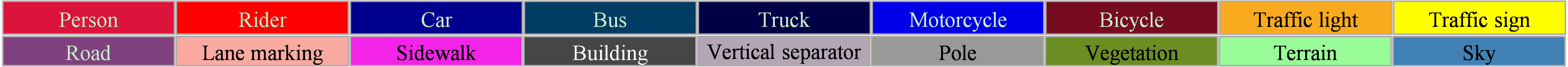}}
		\caption{Examples of results on the test set of SVScape. The results of front, rear, left and right view are displayed in (a). The first two rows show raw image and ground truth, and the following four rows show the results produced by different models. The last row show the improvement/error map which denotes the pixels misclassified by the proposed method in red and the pixels that are misclassified by the base model ERFNet but correctly predicted by the proposed method in green. The color code is listed in (b).}
		\label{exp_real_compare}
	\end{figure*}

	We compare the proposed approach with fine-tuned FCN-VGG16, ENet, and ERFNet on the test set of SVScape. These models are fine-tuned from the pretrained weights on Cityscapes dataset. Table~\ref{exp_compare} shows per-class accuracy results. The proposed approach achieves the top accuracy on all the 18 evaluated classes. The challenging classes (e.g. bus, motorcycle, person, traffic light, traffic sign, pole) improves much more than the general classes (e.g. road, sky, tree, building). The challenging classes have significantly less training samplings than the general classes in the dataset. It suggests our approach which combines a small number of real-world surround view images and a large number of transformed images is effective to improve the performance when the training data is limited. Some examples are shown in Fig.~\ref{exp_real_compare}.
	
\begin{figure*}[htbp]
		\centering
		\includegraphics[width=0.9\textwidth]{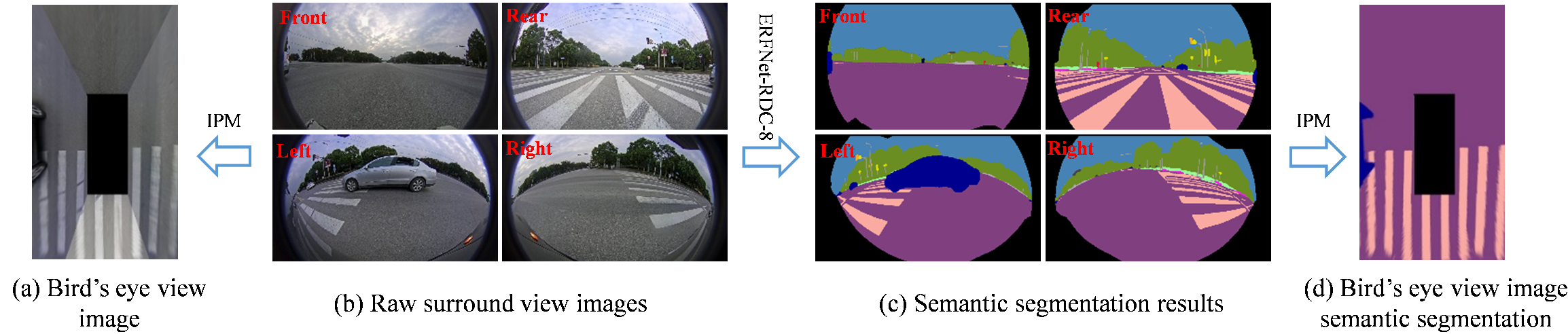}
		\caption{The bird's eye view image semantic segmentation by mapping segmentation results of raw surround view images to bird's eye view plane.}
		\label{fig: Fig. 11}
	\end{figure*}

	The road scene semantic segmentation results on raw surround view images can be used as an information source for other tasks, e.g., visual semantic odometry\cite{lianos2018vso}, Stixel World\cite{schneider2016semantic,deng2016stixel}  and collision warning\cite{wang2015vehicle}. For example, in order to get semantic segmentation results on a bird's eye view image, we can first compute semantic segmentation results on raw surround view images and then map the results to the bird's eye view plane using Inverse Perspective Mapping (IPM), as illustrated in Fig.~\ref{fig: Fig. 11}.
	
	Table~\ref{tab: time} reports the forward pass time of ERFNet and ERFNet-RDC-8 on a single GTX 1080Ti. ERFNet-RDC-8 remains efficient, taking $2$ ms more than the ERFNet that can run at several FPS on an embedded GPU\cite{romera2017erfnet}. The time values reported in Table~\ref{tab: time} include data transfer time from CPU to GPU and the processing time on the GPU, but do not cover the preprocessing time on the CPU and data transfer time from GPU to CPU.

	\begin{table*}[t]
	\centering
	\renewcommand{\arraystretch}{1.3}
	\caption{Per-Class IoU (\%) on the Test Set of SVScape}
	\label{exp_compare}
	\begin{tabu} to 1.0\textwidth{X[3.8,c]|X[1,c]|X[1,c]|X[1,c]|X[1,c]|X[1,c]|X[c]|X[c]|X[c]|X[c]|X[c]|X[c]|X[c]|X[c]|X[c]|X[c]|X[c]|X[c]|X[c]|X[1,c]}
		\hline
		Network & Roa & LMa & TLi & TSi & Pol & VSe & Sid & Per & Rid & Car & Tru & Bus & Mot & Bic & Bui & Tre & Ter & Sky & mIoU\\
		\hline
		\hline
		FCN-VGG16 & \tc{92.3}{70.6}{41.2}{40.4}{28.0}{52.4}{59.9}{52.2}{47.0}
		& \tc{81.9}{82.1}{74.2}{44.5}{59.0}{71.2}{85.7}{58.0}{92.3}
		& $62.9$ \\
		ENet & \tc{93.2}{79.7}{30.8}{40.3}{41.3}{42.7}{61.3}{53.2}{45.5} & \tc{81.9}{81.5}{42.1}{43.2}{57.6}{71.2}{86.8}{64.6}{92.9} & $61.7$\\
		ERFNet & \tc{94.1}{80.5}{50.7}{52.3}{47.3}{57.3}{65.6}{58.2}{52.6}&\tc{82.8}{85.7}{62.6}{49.2}{63.3}{74.4}{88.2}{65.8}{93.9} & $68.0$ \\
		\textbf{Proposed} & \TC{95.1}{83.9}{57.9}{59.7}{55.5}{64.3}{73.4}{66.2}{60.1} & \TC{88.7}{89.3}{82.7}{57.9}{68.6}{77.0}{89.5
		}{70.7}{94.6} & \boldmath{$74.2$}\\
		\hline
	\end{tabu}
\end{table*}	

	\section{Conclusion}
	This paper provides a solution for CNN-based surrounding environment perception using surround view cameras. First, the Restricted Deformable Convolution (RDC) is proposed to enhance the transformation modeling capability of CNNs, so that the net can handle the images with large distortions. Second, in order to enrich surround view training data which are lacking, the zoom augmentation method is proposed to transform conventional images to fisheye images. Two existing complementary datasets are transformed using this method. Finally, an RDC based semantic segmentation model is trained for real-world surround view images through a multi-task learning architecture with the approaches of AdaBN and HLW. Experiments have shown that the RDC based network can effectively handle fisheye images. And the proposed solution was successfully implemented for road scene semantic segmentation using surround view cameras. 
	
	RDC has a good ability to model geometric transformations and is less prone to saturation. Deformable convolution shows a better ability of modeling geometric transformations if only applied to the last few convolutional layers. As future work, RDC and deformable convolution should be combined in one network to further enhance the CNNs' transformation modeling ability. Future work also needs to incorporate weakly supervised or other domain adaptation methods to further improve the performance on real surround view images.
	
	\ifCLASSOPTIONcaptionsoff
	\newpage
	\fi

	
	
	\bibliographystyle{bibtex/IEEEtran}
	\bibliography{bibtex/IEEEabrv_lty,bibtex/Fisheye,bibtex/SemanticSegmentation,bibtex/DeepLearning,bibtex/DomainAdaptation,bibtex/CyberC3}

	%
	\begin{IEEEbiography}[{\includegraphics[width=1in,height=1.25in,clip,keepaspectratio]{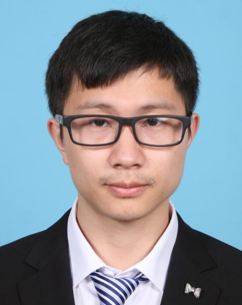}}]{Liuyuan Deng}
		received the B.S. degree in electrical engineering and automation from University of Electronic Science and Technology of China, Chengdu, China, in 2013. 
		He is currently working toward the Ph.D. degree in control science and engineering with Shanghai Jiao Tong University, Shanghai, China. His research interests include computer vision, deep learning, semantic scene understanding and visual localization for intelligent driving.
	\end{IEEEbiography}
	
	\begin{IEEEbiography}[{\includegraphics[width=1in,height=1.25in,clip,keepaspectratio]{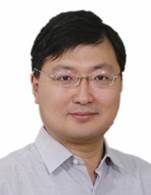}}]{Ming Yang}
		received the master’s and Ph.D. degrees from Tsinghua University, Beijing, China, in 1999 and 2003, respectively. He is currently the Full Tenure Professor at Shanghai Jiao Tong University, director of the Department of Automation, and the deputy director of the Innovation Center of Intelligent Connected Vehicles. He has been working in the field of intelligent vehicles for more than 20 years.
	\end{IEEEbiography}
	
	\begin{IEEEbiography}[{\includegraphics[width=1in,height=1.25in,clip,keepaspectratio]{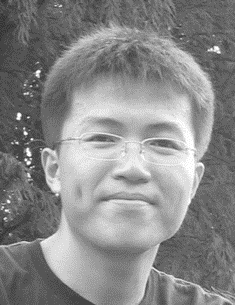}}]{Hao Li}
		is with SJTU-ParisTech Elite Institute of Technology and also with Department of Automation, Shanghai Jiao Tong University, Shanghai, China. Hao Li received the Ph.D. degree in real-time informatics, robotics, and automatics from the Robotics Center of MINES ParisTech, Paris, France, in 2012. He received the master degree and bachelor degree from Department of Automation, SJTU, respectively in 2009 and 2006. His research interests are in artificial intelligence, estimation theory, multisensor data fusion, computer vision, and intelligent vehicle systems.
	\end{IEEEbiography}
	
	\begin{IEEEbiography}[{\includegraphics[width=1in,height=1.25in,clip,keepaspectratio]{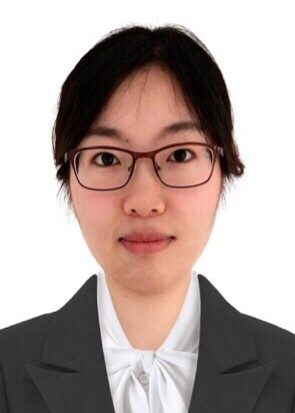}}]{Tianyi Li}
		received the B.S. degree in Automation from Shanghai Jiao Tong University, Shanghai, China, in 2014, where she is currently working toward the Ph.D. degree in control science and engineering.
	
		Her current research interests include autonomous driving systems, time series analysis, information fusion theories, precise positioning and localization, and vehicle dynamics, with an emphasis on design and implementation of vehicle self-localization systems.

	\end{IEEEbiography}
	
	\begin{IEEEbiography}[{\includegraphics[width=1in,height=1.25in,clip,keepaspectratio]{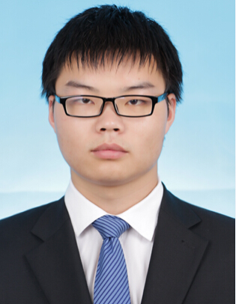}}]{Bing Hu}
		was born in Henan province, China, in 1994. He received the B.S. degree in Automation from SiChuan University, Chengdu, China, in 2015. 
		
		He is currently working toward M.S. degree in control science and engineering with Shanghai Jiao Tong University, Shanghai, China.
		His research interests include computer vision, localization, navigation, SLAM and path tracking in the applications of intelligent vehicle.
	\end{IEEEbiography}
	
	\begin{IEEEbiography}[{\includegraphics[width=1in,height=1.25in,clip,keepaspectratio]{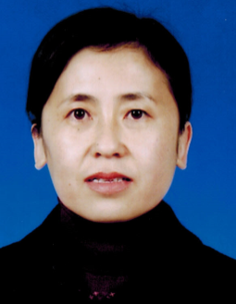}}]{Chunxiang Wang}
		received the Ph.D. degree in Mechanical Engineering from Harbin Institute of Technology, China, in 1999. She is currently an associate professor with the Department of Automation, Shanghai Jiao Tong University, Shanghai, China. Her research interests include autonomous driving, assistant driving, and mobile robots, etc. 
		She has been working in the field of intelligent vehicles for more than 10 years and participated in several related research projects, such as European CyberC3 project, ITER transfer cask project, etc.
	\end{IEEEbiography}
	
	

\end{document}